\documentclass{article}
\usepackage{ijcai24}

\usepackage{times}
\usepackage{soul}
\usepackage{url}
\usepackage[hidelinks]{hyperref}
\usepackage[utf8]{inputenc}
\usepackage[small]{caption}
\usepackage{graphicx}
\usepackage{amsmath}
\usepackage{amsthm}
\usepackage{booktabs}
\usepackage{algorithm}
\usepackage{algorithmic}
\usepackage[switch]{lineno}
\usepackage{amsfonts}

\title{On mitigating stability-plasticity dilemma in CLIP-guided image morphing \\ via geodesic distillation loss}

\author{
Yeongtak Oh$^1$
\and
Saehyung Lee$^1$\and
Uiwon Hwang$^3$*\And
Sungroh Yoon$^{1,2}$*
\affiliations
$^1$Department of Electrical and Computer Engineering, Seoul National University\\
$^2$Interdisciplinary Program in Artificial Intelligence, Seoul National University \\
$^3$Division of Digital Healthcare, Yonsei University\\
*: Corresponding authors \\
\emails
\{dualism9306, halo8218, sryoon\}@snu.ac.kr,
uiwon.hwang@yonsei.ac.kr}

\begin{document}

\maketitle

\begin{abstract}
    Large-scale language-vision pre-training models, such as CLIP, have achieved remarkable text-guided image morphing results by leveraging several unconditional generative models. However, existing CLIP-guided image morphing methods encounter difficulties when morphing photorealistic images. Specifically, existing guidance fails to provide detailed explanations of the morphing regions within the image, leading to misguidance. In this paper, we observed that such misguidance could be effectively mitigated by simply using a proper regularization loss. Our approach comprises two key components: 1) a geodesic cosine similarity loss that minimizes inter-modality features (i.e., image and text) on a projected subspace of CLIP space, and 2) a latent regularization loss that minimizes intra-modality features (i.e., image and image) on the image manifold. By replacing the na\"ive directional CLIP loss in a drop-in replacement manner, our method achieves superior morphing results on both images and videos for various benchmarks, including CLIP-inversion.
\end{abstract}
\section{Introduction}
 Nowadays, deep learning-based text-guided image morphing has been showing unprecedented high qualities in many real-world applications, such as image editing~\cite{patashnik2021styleclip,kim2022diffusionclip}, and style transfer~\cite{kwon2022clipstyler,huang2022diffstyler}. Especially, text-guided image morphing only uses text to give guidance on the given images and does not require any additional target images to guide how to morph.
 
\begin{figure}[t!]
\includegraphics[width=\linewidth]{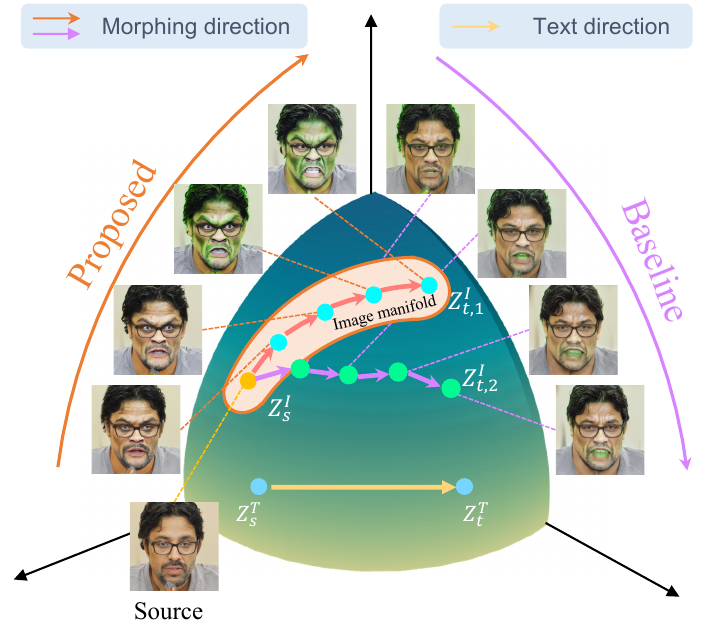} 
\caption{The visualization represents the CLIP space, where image and text features are $L_{2}$-normalized, illustrating an example of morphing from `human' to `hulk'. In CLIP-guided image morphing, $Z^{I}_{s}$ continuously transforms into $Z^{I}_{t}$ by following the text guidance of $Z^{T}_{s}$ to $Z^{T}_{t}$. Here, $Z^{I}$ and $Z^{T}$ denote image and text features, respectively. In our proposed method, the feature of a morphed image is represented by $Z^{I}_{t,1}$, whereas the baseline method employs $Z^{I}_{t,2}$. Specifically, our approach guides the morphing process along the image manifold, resulting in more photorealistic morphed images.}
\label{fig:main}
\end{figure}
Utilizing contrastive language-image pre-training models such as CLIP\footnote{In this paper, we refer to such multi-modal large-scale pre-trained models as CLIP.}~\cite{radford2021learning} is becoming a {\it de facto} choice for text-guided image morphing. This can be achieved by fine-tuning pre-trained generative models like StyleGAN~\cite{gal2022stylegan} and DDPM~\cite{kim2022diffusionclip}, or by explicitly morphing the given images~\cite{kwon2022clipstyler}. Previous work on CLIP-guided image morphing commonly focuses on minimizing spherical distances~\cite{crowson2022vqgan,sauer2023stylegan} or directional CLIP loss~\cite{gal2022stylegan,patashnik2021styleclip,kwon2022clipstyler,song2022clipvg,bar2022text2live,chefer2022image,nitzan2023domain} between normalized image and text features in CLIP space \cite{tevet2022motionclip}. As depicted in Fig. \ref{fig:main}, the textual guidance can be easily obtained in Euclidean space by subtracting the features of the source and target texts in CLIP space \cite{gal2022stylegan}. 

 However, such text-based guidance does not provide detailed information on the specific morphing directions of the source images ({\it e.g.}, the transition from human to hulk). Morphing the source images solely based on such text guidance in CLIP space can result in target images that deviate significantly from the image manifold \cite{zhu2016generative} of the source images. To address this issue, previous methods have tried to alleviate such intrinsic misguidance by imposing a threshold for positive cosine similarity \cite{kwon2022clipstyler}, controlling domain-specific hyperparameters \cite{gal2022stylegan}, and enabling layered edits that combine the edited RGBA layer with the inputs \cite{bar2022text2live}. However, such image modulation requires extensive manual tuning to find optimal hyperparameters or fine-tune the model for obtaining suitable target images. 
 
 In contrast to existing methods, we focus on ensuring that CLIP-guided image morphing proceeds along the CLIP space without deviating from the image manifold. To achieve this, we revisit the {\it stability and plasticity} (SP) dilemma, a prevalent problem in the field of continual learning that is related to the challenge of overcoming catastrophic forgetting \cite{kirkpatrick2017overcoming,li2017learning,hou2019learning,simon2021learning,rebuffi2017icarl,li2017learning}.
 
 That is, the more restrictions there are on learning, the more the model hesitates to learn the new incoming information. Conversely, the more restrictions on memorization, the more the model forgets the previously learned information. Interestingly, in CLIP-guided morphing, we observed that a similar SP dilemma commonly exists in previous methods as follows: 1) drastically morph the given images, leading the morphed images to forget the detailed attributes of the source images, or 2) morph the given images scarcely, which cannot explicitly transform the given images following text guidance. We noticed that this misguidance stems from disregarding the image manifold. To overcome such difficulties, our approach aims to find a compromise morphing direction that preserves essential attributes while effectively following text guidance.
 
 A geodesic distillation loss introduced by \cite{simon2021learning} projects the features from different models onto an intermediate subspace. By minimizing distances in this subspace, the SP dilemma is effectively alleviated, allowing gradual learning without forgetting important features along the geodesic path. Thus, we propose a novel perspective on CLIP-guided image morphing that leverages the advantages of geodesic distillation loss to consider the geodesic path within the CLIP space's image manifold.
 
 Our method minimizes differences between inter-modality ({\it i.e}, image and text) and intra-modality ({\it i.e.}, consecutive images) features, while considering the geodesic path. By employing geodesic cosine similarity in the subspace of the CLIP space, our approach enables photorealistic morphing along the image manifold. For instance, for the case of ‘human’ to ‘hulk’ morphing, our proposed method shows better morphing results compared to the previous method, as shown in Fig. \ref{fig:main}. While morphing the image, the previous baseline method misguides the direction to morph the target images when sophisticated tunings for the unseen domains are absent. In contrast, in the same setting, the proposed method yields significantly better photorealistic morphing results. The benchmark used is StyleGAN-NADA \cite{gal2022stylegan}. 
 
 To the best of our knowledge, our proposed approach is the first to revisit the SP dilemma in the context of CLIP-guided image morphing while considering the manifold structure of CLIP. Through extensive experiments, we consistently demonstrate the superiority of our method by simply replacing the previous directional CLIP loss in a drop-in-replacement manner. 
 The summarization of this paper is as follows. 
\begin{itemize}
    \item In the context of CLIP-guided image morphing, we observed that existing methods are often guided to generate non-photorealistic images caused by the inherent challenges associated with the SP dilemma.
    \item To address such misguidance, we propose a novel approach that effectively morphs the image by faithfully reflecting the text guidance. Motivated by \cite{simon2021learning}, our method involves regularization of the morphing directions within the image manifold by following the geodesic path on the feature-dependent subspace of the CLIP space.
    \item We corroborate that the proposed method consistently produces photorealistic image morphing results on several benchmarks, including StyleGAN-NADA and Text2Live.
    \item Additionally, we design a CLIP inversion method that does not require pre-trained generators to morph the image and show the superiority of the proposed method.
\end{itemize}
\section{Preliminaries}
\subsection{Contrastive Language-Vision Pre-training Model}
Large-scale pre-trained language-image models like CLIP \cite{radford2021learning}, OpenCLIP \cite{cherti2022reproducible}, and Align \cite{jia2021scaling} have exhibited remarkable robustness to natural distribution shifts \cite{fang2022data}. These models are trained on extensive image and unstructured text pairs sourced from the web. Image and text encoders of CLIP are jointly trained by minimizing the InfoNCE \cite{oord2018representation} loss, which minimizes the distance between the two modalities (\textit{i.e.}, image and text). As a result, CLIP can align the input image-text pairs for zero-shot image classification \cite{zhou2022conditional,li2022masked,liang2022mind}, text-guided image generation \cite{rombach2022high,ramesh2022hierarchical}, and text-guided image morphing \cite{gal2022stylegan,bar2022text2live,kwon2022clipstyler,kim2022diffusionclip}. In this paper, different from the text-guided image generation, which only utilizes a text encoder of CLIP for training generative models, we utilize both image and text encoders of CLIP for image morphing.
\begin{figure*}[t!]
    \includegraphics[width=\linewidth]{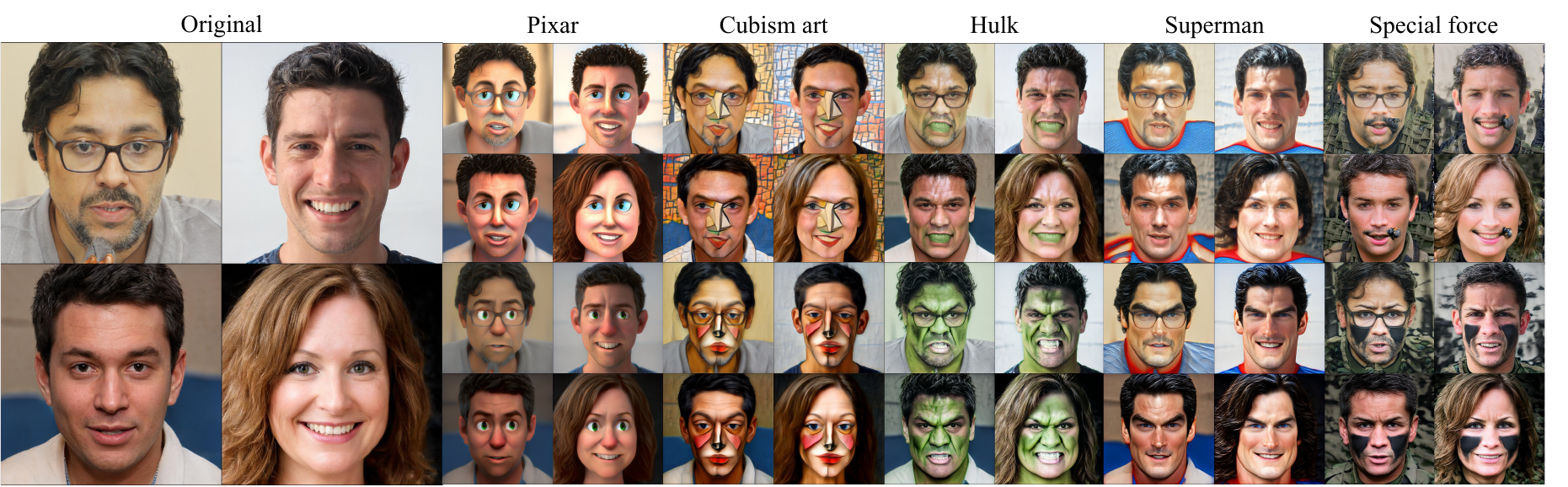}  \caption{Results of the CLIP-guided image morphing. Original images are generated from StyleGAN pre-trained with FFHQ dataset. The first row is the result of the baseline method, and the second row is the result of the proposed method.}
    \label{fig:style_1}
\end{figure*}
\subsection{Text-guided image morphing via CLIP}
Conventionally, image morphing \cite{lee1996image} involves a smooth transformation from one image to another. Through such image metamorphosis, this process generates a sequence of intermediary images that gradually transition into the target images. In contrast to image-to-image morphing, text-guided image morphing allows for the manipulation of source images using specific concepts (\textit{i.e.} prompts) without the need for target images. \\
\indent To morph a given image, the directional CLIP loss \cite{gal2022stylegan,bar2022text2live,kim2022diffusionclip} in Eq. (\ref{eq:positive_sim}) or the squared spherical distance \cite{crowson2022vqgan,sauer2023stylegan} in Eq. (\ref{eq:sphere_sim}) are frequently used. 
\begin{equation}
    \mathcal{L}^{\textrm{dir}}_{\textrm{CLIP}}=1-\textrm{cos}(\Delta{z_{I}},\Delta{z_{T}})=1-\frac{\Delta{z_{I}}\cdot\Delta{z_{T}}}{|\Delta{z_{I}}|\cdot|\Delta{z_{T}}|}
\label{eq:positive_sim}
\end{equation}
where $\Delta$ is the direction from source to target, and $\Delta{z_{I}}=E_{I}({x^{I}_{\textrm{target}})-E_{I}(x^{I}_{\textrm{source}}})$, $\Delta{z_{T}}=E_{T}({x^{T}_{\textrm{target}})-E_{T}(x^{T}_{\textrm{source}}})$. Here, $x^{I}$, $x^{T}$ denote the image and texts, respectively. $E_{I}$ and $E_{T}$ are the CLIP image and text encoders, respectively. For the case of fine-tuning the pre-trained generator, $\Delta{z_{I}}=E_{I}(G_{\textrm{train}}({x_{I, \textrm{t}}))-E_{I}(G_{\textrm{frozen}}(x_{I, \textrm{s}}}))$).
\begin{equation}
    \mathcal{L}^{\textrm{dir}}_{\textrm{sphere}}=1-\textrm{arccos}^{2}(\frac{\Delta{z_{I}}\cdot\Delta{z_{T}}}{|\Delta{z_{I}}|\cdot|\Delta{z_{T}|}})
\label{eq:sphere_sim}
\end{equation}
\indent In this paper, we utilized StyleGAN-NADA and Text2Live as benchmarks and demonstrated the effectiveness of our proposed method compared to the benchmarks even without altering any hyperparameters.
\section{Related works}
\indent Based on our observations that existing CLIP guidance induces SP dilemma, we aimed to improve the CLIP guidance. In the domain of continual learning, to mitigate the SP dilemma, cosine normalization of features \cite{hou2019learning} is introduced to address the class imbalance problem. Simon et al. \cite{simon2021learning} further improved upon \cite{hou2019learning}'s approach by proposing a geodesic distillation loss within an intermediate subspace formed by two distinct models, \textit{i.e.}, learned from the previous and current tasks. \\
\indent Similar to our insights, \cite{zhou2022clip} revealed that a full-dimensional CLIP space fails to effectively capture useful visual information, while an emotional subspace better captures changes in facial attributes. Additional domain modulation operations \cite{alanov2022hyperdomainnet} are introduced to address the multi-domain adaptation problem in GANs. Next, in \cite{nitzan2023domain}, it is demonstrated that a pre-trained generator can harmoniously expand in dormant directions within the latent space and can be linearly expanded using repurposed directions from the base subspace. However, these methods have limitations as they do not consider the manifold of CLIP and rely solely on linearized directions in the latent space. Hence, there is still a lack of proper CLIP guidance design, and the reasons why image morphing should be considered within the subspace of CLIP have not been investigated.
\section{Mitigating SP dilemma in Morphing:\\ Geodesic path in CLIP}
\subsection{Interpret CLIP-guided image morphing through the lens of continual learning}
\indent In this section, we elucidate that our approach is significantly different from the goal of continual learning. Specifically, the work described in \cite{simon2021learning} was primarily designed for class-incremental learning, which is specific to classification tasks. In contrast, our research deals with multi-modal data and aims to gradually morph the image following the text guidance. Recognizing that CLIP functions as a cosine classifier for normalized features of different modalities, we discerned the potential to apply a similar intuition to enhance CLIP guidance. We present a novel approach that leverages the SP dilemma to enhance image morphing and achieve more photorealistic results. We focused on our findings that maximizing cosine similarity in the full-dimensional CLIP space (\textit{e.g.}, 512 for ViT-B/32 \cite{dosovitskiy2020image}) would readily lead to misguided image morphing that significantly morphs the detail attributes or rarely morphs the crucial attributes of source images. To address this issue, we propose conducting CLIP-guided image morphing in a low-dimensional subspace of CLIP. 
\subsection{Analytic derivations of geodesic flow among different models} 
\indent Following \cite{simon2021learning}, they enforced consistency along the \textit{geodesic flow} on the Grassmann manifold. Grassmann manifold \cite{bendokat2020grassmann} is widely used to cope with problems such as low-rank matrix optimization. This approach enables gradual changes of the new model from the source model by projecting each important knowledge onto the intermediate feature subspace. In our work, we extend the notion of the geodesic flow to connect two different features (\textit{i.e.}, image-text or image-image) in the CLIP space for CLIP-guided image morphing. \\
\indent Let $z_{t}$ and $z_{t+1}$ be features of the model at the $t^{\text{th}}$ and $(t+1)^{\text{th}}$ learning phases, respectively. Consider a metric space composed of two embedded features $z$ and $\hat{z}$ within their intermediate subspace $Q$. The inner product in this space can be defined as $z^{T} Q \hat{z}$. Then the geodesic flow $\Pi:\nu\in[0,1]\rightarrow\Pi(\nu)\in\mathcal{G}(N,D)$ is defined between the orthonormal basis of $P_{t}$ and $P_{t+1}$ as follows:
\begin{equation}
    \Pi(\nu) = 
    \begin{bmatrix} P_{t} & R \end{bmatrix}
    \begin{bmatrix} U_{1}\Gamma(\nu) \\ -U_{2}\Sigma(\nu) \end{bmatrix}
\label{eq:geodesic_flow}
\end{equation}
where $R\in\mathbb{R}^{D\times(D-N)}$ is the orthogonal complement of $P_{t}$, and $U_{1}$ and $U_{2}$ are orthonormal matrices satisfying $P^{T}_{t}P_{t+1}=U_{1}\Gamma V^{T}$, and $R^{T}_{t}P_{t+1}=U_{2}\Sigma V^{T}$. Note that, principal component analysis (PCA) is used to obtain $P_{t}$ and $P_{t+1}$ to project the features $z_{t}$ and $z_{t+1}$ into a low dimensional space. Furthermore, the orthogonal complement and the diagonal elements could be calculated using a singular value decomposition (SVD) \cite{van1976generalizing} algorithm. \\
\begin{equation}
Q=\int_{0}^{1}  \Pi(\nu)^{T} \Pi(\nu) \,d\nu
\label{eq:q_matrix}
\end{equation}
\indent In Eq. (\ref{eq:q_matrix}), the integration of the inner product $Q$ is defined as a positive semi-definite matrix with the size of $D\times D$, which denotes an intermediate subspace on the Grassmann manifold. Analytic derivations of the geodesic flow $\Pi(\nu)$ and the matrix $Q$ are noted in \cite{simon2021learning}. The total geodesic distillation loss on an intermediate space can be expressed as follows:
\begin{equation}
    \mathcal{L}^{\textrm{Geo}} = 1 - \frac{z_{t}Qz_{t+1}}{||Q^{1/2}z_{t}||\cdot||Q^{1/2}z_{t+1}||} \label{eq_original}
\end{equation}
As reported in \cite{simon2021learning}, if $P_{t}$ and $P_{t+1}$ are identical, Eq. (\ref{eq_original}) is equivalent to the na\"ive cosine similarity loss.
\subsection{CLIP-guided morphing via geodesic distillation loss}
\indent In this section, we explain our proposed approach, both considering the multi-modality and uni-modality regularization losses as following subsections. 
\subsubsection{Inter-modality consistency (IMC) loss}
To maximize the cosine similarity between two distinct image and text features in the feature-dependent subspace of CLIP, we define IMC loss as follows: 
\begin{equation}
    \mathcal{L}^{\textrm{Inter}}_{\textrm{Cons}} = 1 - \frac{\Delta{z^{I_{i}}}Q_{\textrm{Inter}}\Delta{z^{T}}}{||Q_{\textrm{Inter}}^{1/2}\Delta{z^{I_{i}}}||\cdot||Q_{\textrm{Inter}}^{1/2}\Delta{z^{T}||}} \label{eq_geodesic1}
\end{equation}
where, $\Delta{z^{I_{i}}}=\frac{E_{I}(I_{i})-E_{I}(I_{s})}{|E_{I}(I_{i})-E_{I}(I_{s})|}$, $\Delta{z^{T}}=\frac{E_T(T_{t})-E_T(T_{s})}{|E_T(T_{t})-E_T(T_{s})|}$. $i$ is the timestep where $i\in[1,2,\cdots,\textrm{T}]$. $I_{s}$ represents the source image, and $(T_{t}, T_{s})$ represent the target and source text, respectively. This loss term describes discrepancies between the image features and text features within CLIP space. Consequently, by minimizing IMC loss, modality mismatches between the provided image and text features within the full-dimensional CLIP space are gradually alleviated by projecting them onto a lower-dimensional subspace.
\subsubsection{Intra-modality regularization (IMR) loss}
To regularize the morphing direction between two consecutive images, we describe IMR loss as follows:
\begin{equation}
    \mathcal{L}^{\textrm{Intra}}_{\textrm{Reg}} = 1 - \frac{z^{I_{i-1}}Q_{\textrm{Intra}}z^{I_{i}}}{||Q_{\textrm{Intra}}^{1/2}z^{I_{i-1}}||\cdot||Q_{\textrm{Intra}}^{1/2}z^{I_{i}}||} \label{eq_geodesic2}
\end{equation}
where, $z^{I_{i}}=\frac{E_{I}(I_{i})}{|E_{I}(I_{i})|}$, $i\in[1,2,\cdots,\textrm{T}]$, and $I_{0}=I_{s}$. This loss term represents the differences of image features in the subspace of CLIP, and by minimizing IMR loss, images are guided to gradually morph following the smoothed geodesic path without deviating from the image manifold. Note, this consecutive regularization in-between image features is somewhat aligned with the aim of continual learning. \\
\indent Thus, to facilitate the CLIP guidance by considering two losses, our total loss term is as follows:
\begin{equation}
    \mathcal{L}^{\textrm{Total}}=\mathcal{L}^{\textrm{Inter}}_{\textrm{Cons}}+\lambda_{1}\mathcal{L}^{\textrm{Intra}}_{\textrm{Reg}}+\lambda_{2}\mathcal{L}_{\textrm{LPIPS}}
     \label{eq_total_loss}
\end{equation}
where $\lambda_{1}$ and $\lambda_{2}$ are set to $1$ and $0.3$, respectively. Here, we considered minimizing the LPIPS loss~\cite{zhang2018unreasonable} to significantly enhance the visual quality and achieve more photorealistic outcomes. The comprehensive ablation studies of the employed losses are illustrated in Fig. \ref{fig:style_ablation}. We utilize this loss, denoted as $\mathcal{L}^{\text{Total}}$, for our proposed loss. This total loss represents an augmented version of the commonly used directional CLIP loss. Our proposed CLIP guidance method effectively modifies specified attributes while preserving the essential characteristics of the input images. This approach addresses the inherent challenge of misleading morphing directions, which could otherwise result in the acquisition of undesired attributes or insufficiently morphed features.
\subsection{CLIP inversion}
To demonstrate the effectiveness of our proposed CLIP guidance, we propose CLIP inversion without requiring a pre-trained generator like GAN \cite{karras2020training} or Diffusion \cite{ho2020denoising}. We exploit CLIP inversion to verify that directional CLIP loss induces class-wise catastrophic forgetting of source attributes, which cannot be easily conducted with pre-trained unconditional generative models. We leverage and refine the model-agnostic model inversion \cite{ghiasi2022plug}, which enables image inversion through data augmentation. In contrast to previous studies \cite{ghiasi2022plug,yin2020dreaming}, our CLIP inversion covers multi-modal properties and exploits CLIP's image and text encoders for image morphing. To initiate the image morphing process, initial source images are selected. Subsequently, the selected source images undergo morphing by minimizing the discrepancies between their image and text features, utilizing either the loss defined in Eq. (\ref{eq:positive_sim}) or Eq. (\ref{eq_total_loss}). For CLIP inversion, we utilized various techniques such as DiffAug \cite{zhao2020differentiable}, ensembling method \cite{ghiasi2022plug}, and random perspective, random affine transform were employed to enhance the visual plausibilities of morphed images.

%
\section{Experimental Results}
In the following experiments, we show that our proposed method explicitly enhances the image morphing quality to make it more photorealistic. The subspace dimension is set to $256$ for all experiments, and the CLIP image encoder was set to ViT-B/32 \cite{dosovitskiy2020image}. \\
\indent We provide additional explanations such as CLIP-styler \cite{kwon2022clipstyler}, StyleCLIP \cite{patashnik2021styleclip}, DiffusionCLIP \cite{kim2022diffusionclip} and CLIP-guided latent diffusion models \cite{rombach2022high}, in the Supplementary Material.
For further elucidation and comprehensive details and results, please refer to the Supplementary Material.

\begin{figure}[t!]
   \includegraphics[width=\linewidth]{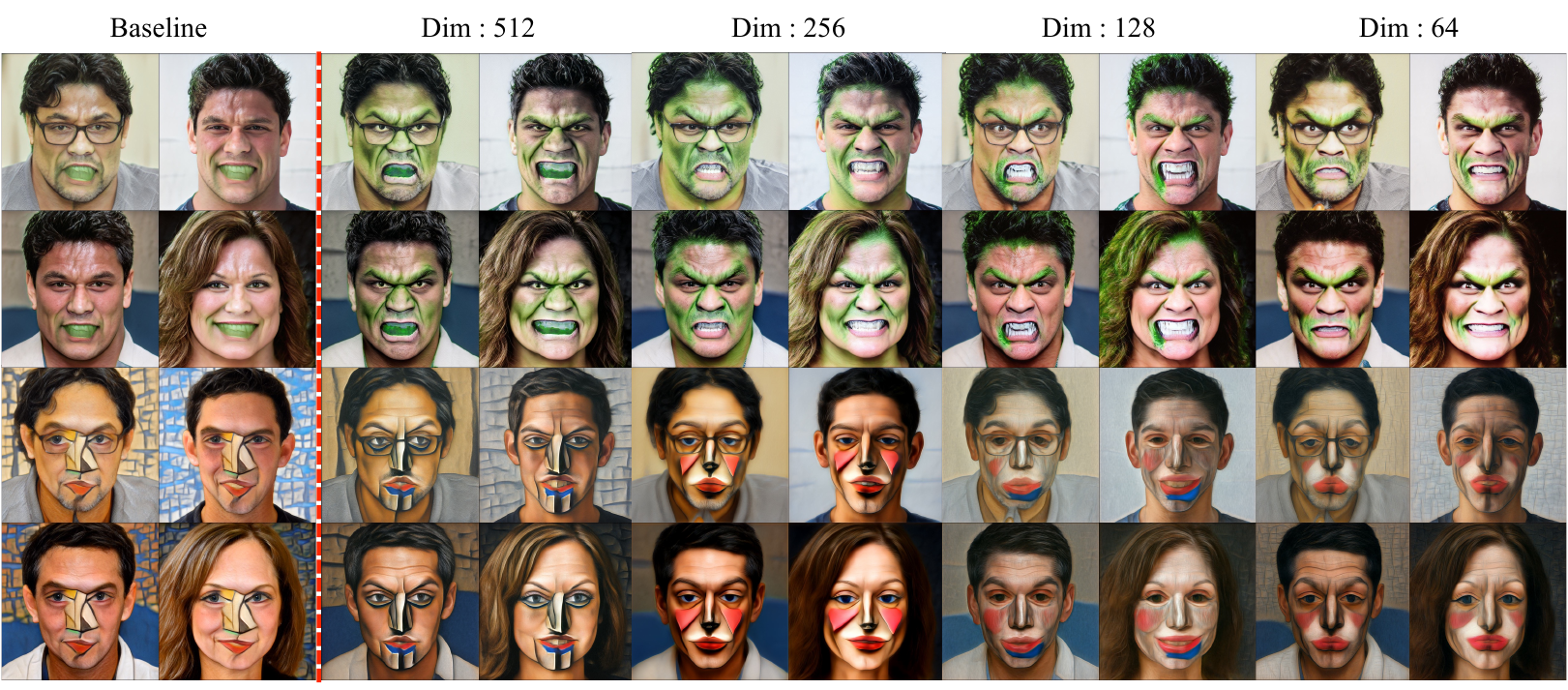} 
    \caption{Dimensional studies to select the optimal value of subspace dimension.}
    \label{fig:dim_study}
\end{figure}
\begin{figure}[t!]
    \centering
    \includegraphics[width=0.8\linewidth]{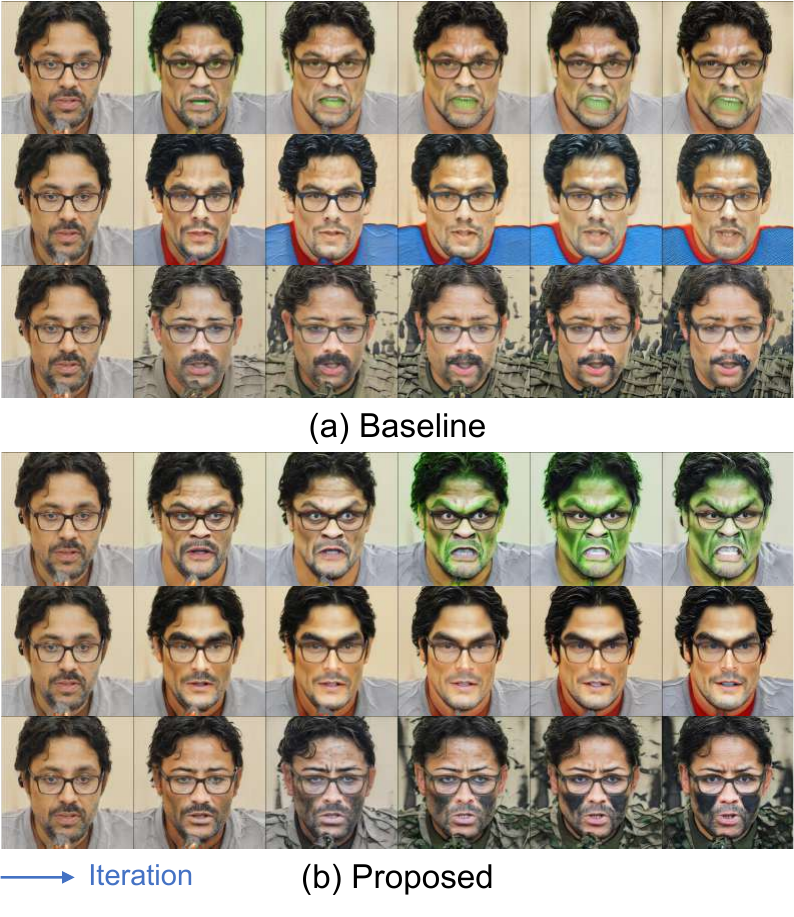} 
    \caption{Continuous image metamorphosis according to the iterations for the cases of `hulk', `superman', and `special forces' with (a) the baseline and (b) our proposed method.}
    \label{fig:steering}
\end{figure}
\begin{figure}[t!]
    \includegraphics[width=\linewidth]{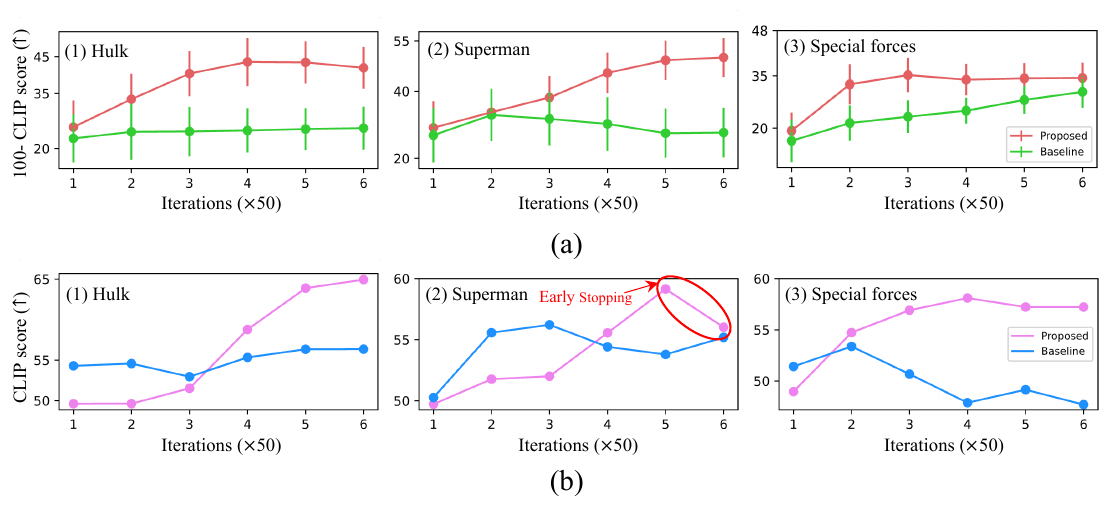} 
    \caption{Visualization of CLIP scores. (a) denotes the extent of image morphing from source images, and (b) denotes the extent of image morphing towards the target image manifold. Our method consistently outperforms the baseline for all of the given prompts and each training iteration.}
    \label{fig:5to7}
\end{figure}
\subsection{Improving StyleGAN-NADA}
\cite{gal2022stylegan} proposed a domain-specific fine-tuning technique for StyleGAN \cite{karras2020training} generators using text guidance. This approach initializes source images with generated images and morphs them according to the provided text guidance in a zero-shot manner. In Fig. \ref{fig:style_1}, we observed that the baseline method tends to generate artifact images, characterized by distorted facial features and unnatural gaze. In contrast, our proposed method consistently outperforms the baseline method while achieving more qualitative morphing, even when using default hyperparameters for all given prompts. Thus, our method shows consistently better results while highly mitigating the hyperparameter reliances. To better emphasize the superiority of our method, we specifically examine the results of morphing, focusing on out-of-domain prompts (\textit{i.e.}, hulk, superman, and special forces) that are not limited to the in-domain morphing directions (\textit{e.g.},  facial changes and gender) of the source data and thus easily lead to misguided morphing directions. \\
\indent Notably, for the Pixar and Cubism art prompts, the baseline method exhibits drastic morphing results that lead to catastrophic forgetting of the given source images. Conversely, for the cases of hulk, superman, and special forces, the baseline method yields negligible changes in attributes on faces and fails to achieve effective results. For instance, as shown in Fig. \ref{fig:style_1}, the baseline struggles to accurately morph Hulk images and results in localized greenish tones on teeth. Conversely, our proposed method generates semantically improved and more realistic morphed results. 
\begin{figure}[t!]
    \includegraphics[width=\linewidth]{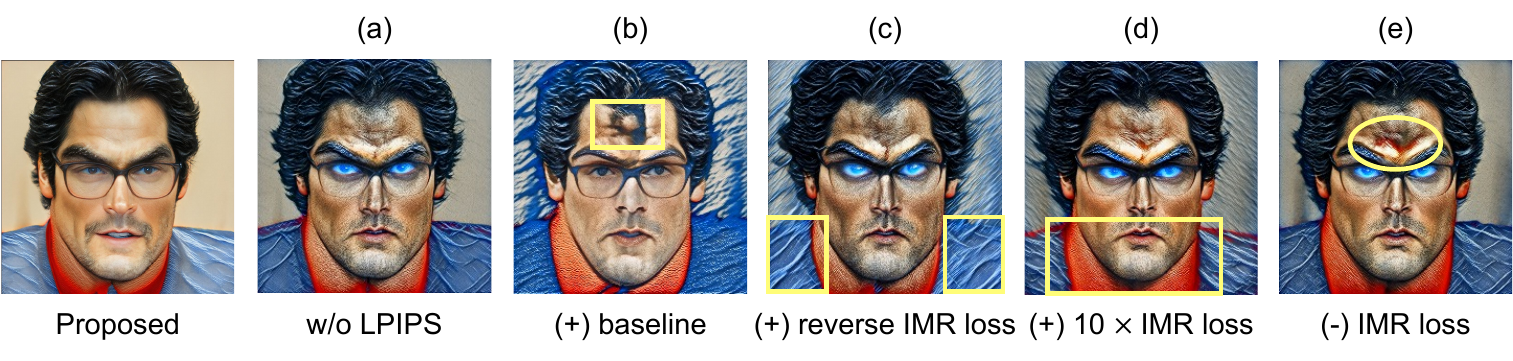}  \caption{Ablation studies of the proposed loss.}
    \label{fig:style_ablation}
\end{figure}
\subsubsection{Dimension study}
\indent We conducted an ablation study to determine the appropriate subspace dimension. As shown in Fig. \ref{fig:dim_study}, when the subspace dimension is significantly low, such as $64$ or $128$, the morphed images do not accurately reflect the text guidance. On the other hand, when using the $512$ dimension, our proposed method exhibits drastic morphing results. Therefore, we selected $256$ as a trade-off for the subspace dimension in our proposed method, as shown in Fig. \ref{fig:dim_study}. In Fig. \ref{fig:steering}, the results indicate that the proposed method outperforms the baseline method by effectively morphing the attributes of the source images at each iteration step. 
\subsubsection{Quality evaluations}
\indent We performed comprehensive quality evaluations. In Fig. \ref{fig:5to7} (a), we evaluated both the baseline and proposed methods using $100$ samples per prompt, for each training iteration. We measured the morphing CLIP score, which is calculated as $100\times(1-\textrm{cos}(E_{I}(x^{src}_{\textrm{image}}), E_{I}(x^{trg}_{\textrm{image}})))$. This score indicates the extent of dissimilarity between the morphed images and the source images. As a result, for all of the given prompts, our method consistently outperforms the baseline in all training iterations. In Fig. \ref{fig:5to7} (b), we measured CLIP scores for the morphed images and target images. This result demonstrates that our method provides unique guidance to reach the target image manifold, which cannot be achieved using a directional CLIP loss. We indicated the early-stopping point for the Superman prompt in the figure. We compare the outcomes of minimizing the directional CLIP loss and our proposed loss in (a) and (b), respectively. 
\begin{figure}[t!]
    \centering
    \includegraphics[width=0.8\linewidth]{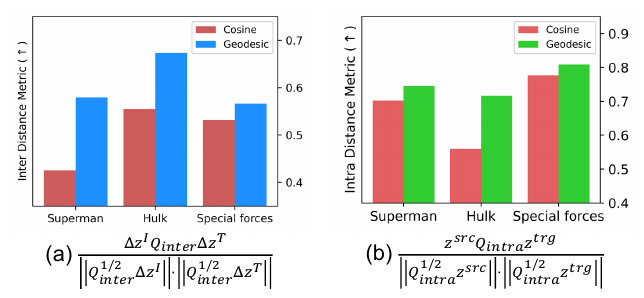}  \caption{Metric distances of the proposed method compared to the baseline.}
    \label{fig:metric_score}
\end{figure}
\subsubsection{Ablation studies}
To evaluate the effectiveness of each proposed loss, we conducted ablation studies whose results are depicted in Fig. \ref{fig:style_ablation}. These results illustrate that our proposed method yields the most photorealistic and high-quality image morphing. Specifically, (a) demonstrates that omitting the LPIPS loss significantly compromises the photorealism of the source images. In scenario (b), incorporating the directional CLIP loss with the proposed loss and minimizing it results in a decline in overall quality. In (c), the guiding directions of the proposed IMR loss are reversed, leading to unrealistic artifacts in the images. Similarly, (d) shows the effects of varying the weighting coefficient of the loss, also resulting in unrealistic artifacts. Finally, (e) indicates that using only the IMC loss leads to the emergence of distinct artifacts.
\subsubsection{Visualization of metric distances}
To demonstrate the claim that \textit{our proposed method morphs the image following the geodesic path within CLIP}, we conducted an analysis of inter and intra $d_{M}$ between source and morphed images. We compared the outcomes of minimizing the directional CLIP loss and our proposed loss in Fig. (a) and (b), respectively. Note that $d_{M}$ is calculated using its normalized features and normalized between $0$ and $1$. In Fig. \ref{fig:metric_score}, we utilize ViT-L/14 CLIP model and its sub-dimension $384$ for evaluation, which was not used for GAN training. The results consistently show that our proposed method achieved higher inter and intra $d_{M}$ than the baseline in all of our experiments. These findings both explain that in the subspace of CLIP, (1) \textit{plasticity, inter-modality}: our guidance effectively aligns image morphing directions with text directions, and (2) \textit{stability, intra-modality}: the features of morphed and source images lie more closely. 
\subsection{Improving Text2Live}
\begin{figure}[t!]
    \includegraphics[width=\linewidth]{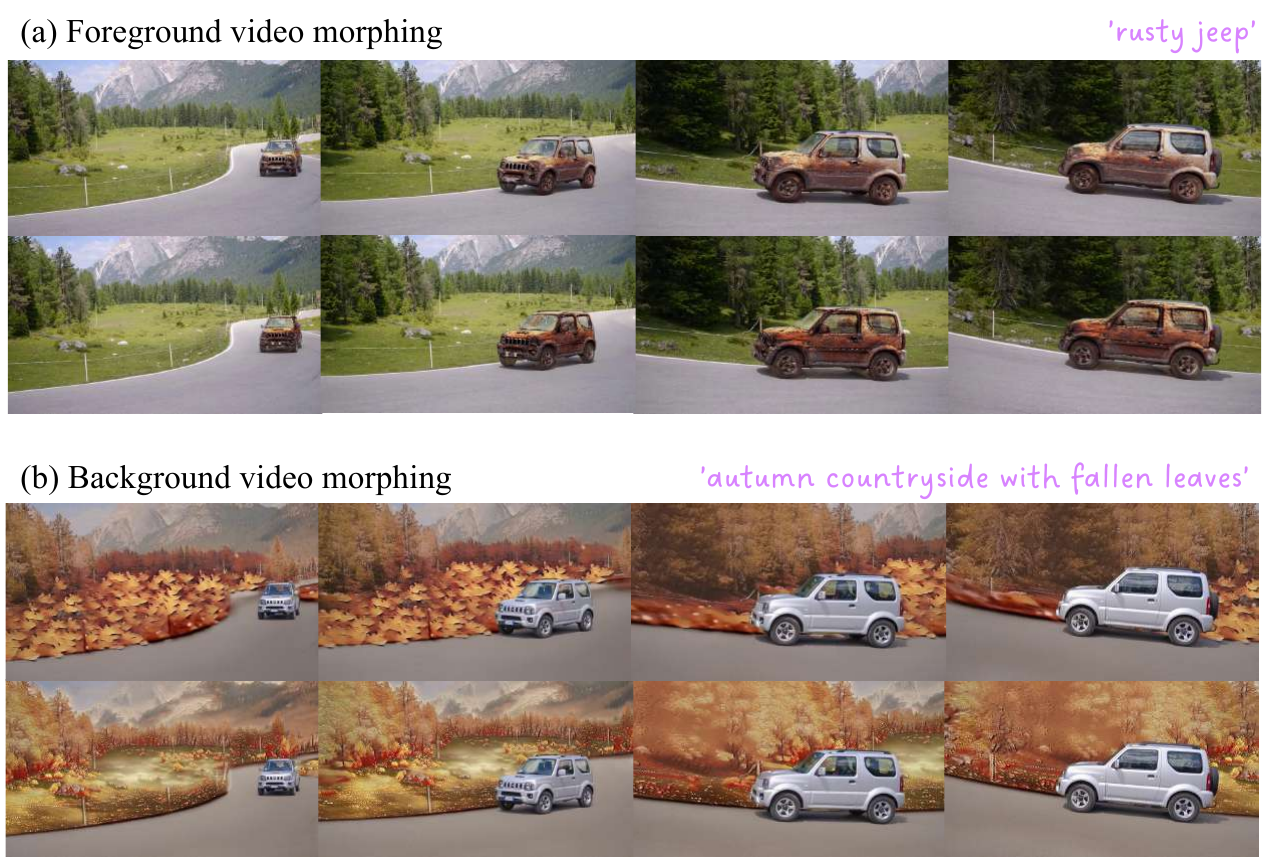}  
    \caption{Results of video morphing for two random prompts. The results of the baseline methods are shown in the first row, and the results of our method are shown in the second row. For all cases, our proposed method shows predominant video morphing results for all of the frames.}
    \label{fig:text2live_video}
\end{figure}

\cite{bar2022text2live} presented a zero-shot manipulation method with newly added visual concepts using texts to augment a given scene or existing objects in a natural and meaningful manner and edit natural images and videos with text guidance. Without loss of generality, also for the video morphing, as depicted in Figure \ref{fig:text2live_video}, our proposed method demonstrates superior results compared to the baseline method in both (a) foreground video morphing and (b) background video morphing experiments. In Fig. \ref{fig:text2live_video} (a), our proposed method effectively transforms the original texture of the selected regions for the `rusty jeep' prompts. In Fig. \ref{fig:text2live_video} (b), more notable differences emerge for the results of our proposed method and the baseline method, primarily because the background regions allow for more room for extensive morphing. While the baseline method drastically alters the original video (\textit{i.e.}, evidenced by the field covered in fallen leaves), our proposed method achieves superior morphing results and maintains photorealism. 
\begin{table}[t!]
    \scriptsize
    \scalebox{1.1}{
    \begin{tabular}{c|c}
        \multicolumn{1}{c}{} \\ \hline
            Source & `A photo of a [cls].' \\ \hline
            Target & `A watercolor painting of a [cls] in the forest.' \\ 
             & `A plush toy of a [cls] in the underwater.' \\
             & `3D Unreal Engine rendering of a [cls] in the rainy day.' \\ 
             & `A pencil drawing of a [cls] on canvas.' \\ 
             & `A character design of a [cls] in the style of pixar.' \\
             & `Oil painting of a [cls] with flowers.' \\ \hline
    \end{tabular}
    }
    \caption{The used prompts for class-wise CLIP inversion experiment. Note, [cls] denotes the specific class names.}
    \label{fig:prompt_table}
\end{table}
\begin{figure}[t!]
    \centering
   \includegraphics[width=\linewidth]{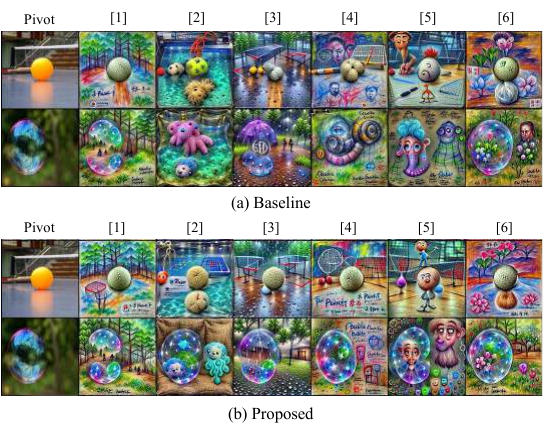} 
    \caption{Results of CLIP inversion by minimizing the loss of (a) Eq. (\ref{eq:positive_sim}) and (b) Eq. (\ref{eq_total_loss}).}
    \label{fig:clip_inversion}
\end{figure}
\begin{figure}[t!]
   \includegraphics[width=\linewidth]{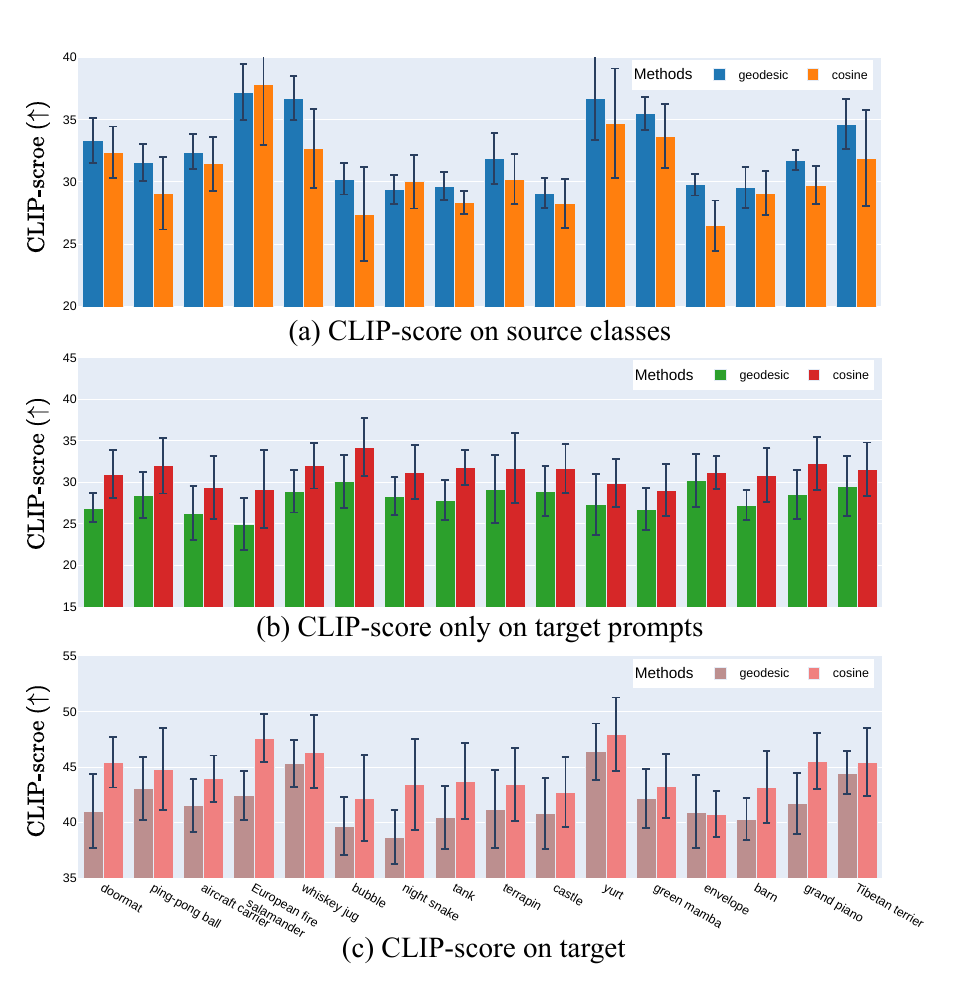} 
    \caption{The results of the CLIP score for morphed images via CLIP inversion with (a) classes without applying prompts, (b) only target prompts without classes, and (c) full target texts including classes.}
    \label{fig:clipinv_total}
\end{figure}
\subsection{Class-wise image morphing via CLIP inversion}
To validate our hypothesis that the directional CLIP loss significantly contributes to the forgetting of source attributes in conditional settings, we conducted a series of class-wise image morphing experiments. In these experiments, we examined how our proposed method preserves important class-wise attributes during text-guided morphing across various prompt scenarios. We used a fixed random seed and randomly selected 16 classes from the ImageNet dataset \cite{deng2009imagenet}. The specific source and target texts employed are detailed in Table \ref{fig:prompt_table}. \\
\indent Fig. \ref{fig:clip_inversion} showcases the results of image morphing utilizing both Eq. (\ref{eq:positive_sim}) and Eq. (\ref{eq_total_loss}). Here, we applied six distinct prompts to source images from the 'ping-pong ball' and 'bubble' classes. In Fig. \ref{fig:clip_inversion}(a), the baseline method morphs the source images according to each text prompt, but it often neglects key attributes (e.g., the shape of the ball and bubble) in several instances. In contrast, our proposed method consistently retains the detailed attributes of the source images for all text prompts. \\
\indent To quantitatively assess our results, in Fig. \ref{fig:clipinv_total}, we evaluated both (a) the preservation of important features from the source images related to the class and (b) the extent of changes of morphed images only related to the target text that is not related to the source classes. This textual decomposition is achieved by dividing the target texts into respective class descriptors (e.g., 'a [cls]') and target prompts (e.g., 'watercolor painting in the forest'), followed by measuring the CLIP scores for each component. These results are quantified using the CLIP score, with the mean and variance displayed in Fig. \ref{fig:clipinv_total}. \\
\indent Interestingly, in Fig. \ref{fig:clipinv_total} (a), our proposed method attains consistently higher CLIP scores specifically related to the given classes. Subsequently, (b) reveals that the baseline method achieves higher CLIP scores for the target texts compared to our proposed method. Lastly, Fig. \ref{fig:clipinv_total} (c) shows that the baseline method scores higher on the target prompts alone, suggesting that it tends to neglect the class information and predominantly aligns the morphing direction with the target prompts. These findings indicate that images morphed using our proposed method more effectively preserve class-specific attributes compared to those generated by the baseline method across all cases. 
\section{Conclusion}
In this paper, we propose a simple yet effective approach while confirming the effectiveness of our proposed method by conducting extensive experiments with several benchmarks, including CLIP-inversion, to improve existing CLIP-guided image morphing. As a result, our proposed method consistently shows predominant photorealistic outcomes and better alleviates the SP dilemma in morphing results for overall settings, with and without pre-trained generators. In future work, we expect our method can be extended to other large-scale CLIP models (\textit{e.g.}, OpenCLIP).
\section{Limitations}
Although our method provides better text-aligned morphing by faithfully following the geodesics in CLIP, we conjecture morphing directions are guided to have several stereotypes of CLIP learned from its training data. Further, not only for our works but also commonly in previous works that exploit zero-shot CLIP, early stopping issues, related to the trade-off between image morphing and photorealism, still remain.
\bibliographystyle{named}
\bibliography{references}

\end{document}


\maketitle
\appendix
\section{Details and Further Results of CLIP Inversion}
Recently, Ghiasi \textit{et. al.} \cite{ghiasi2022plug} proposed a model-agnostic inversion method that can generate a proxy image from an image initialized with random noise by applying various data augmentations, such as jittering, rolling, centering, and zooming. This technique allows pre-trained image classification models, such as ResNets \cite{he2016deep}, Vision transformers \cite{dosovitskiy2020image}, and MLP-mixers \cite{tolstikhin2021mlp}, to invert an image for specific classes learned during pre-training. In other words, model inversion inverts the training data learned by the model onto a condensed image, by capturing the characteristics of the data. Thus, model inversion gradually transforms the image to include data-specific statistics. However, existing methods \cite{ghiasi2022plug} have only demonstrated model inversion for models trained on unimodal (\textit{i.e.} image-only) data, hence their usefulness is limited to inverting data associated with specific classes. Thus, the model trained on data with multi-modality (\textit{i.e.}, image and text) such as CLIP has not yet been considered for the model inversion. \\
\indent To extend the previous model-inversion method considering CLIP, we design a CLIP inversion algorithm. Consequently, the trained web-scaled data statistics of CLIP are inverted onto the images for the given texts. The details are described in Algorithm \ref{alg_clipinv}. Here, $E_{I}$ and $E_{T}$ denote CLIP's image and text encoder, respectively. We used $16$ ensemble images that contain differently applied random augmentations to generate one image without using the pre-trained generator. Note, the more plausible text-guided images could be morphed using the more ensemble images applied with random augmentation, considering empirical risk minimization (ERM). However, although exploiting more ensemble images would enable more photorealistic image morphing, it could raise the out-of-memory (OOM) problem. Thus, by balancing the quality and memory trade-off, we selected $16$ ensemble images to update the image. In Algorithm \ref{alg_clipinv}, the geodesic distillation loss described in our paper or the directional CLIP loss could be utilized for the loss. Note that, we exploited the ViT-B/32 model for CLIP inversion. \\
\indent To enhance the quality of morphed images, we replaced the previously used data augmentations, such as scaling and centering \cite{ghiasi2022plug} for training efficiency. We used Kornia python libraries to apply differentiable random augmentations (\textit{i.e.}, random affine transform, random perspective, and random resize crop) as used in VQGAN-CLIP \cite{crowson2022vqgan}. For CLIP-guided morphing, we set total epochs as $800$, learning rate as $2e^{-4}$, and used a cosine scheduler during training. In our experiments, text guidance is shared in the same batch. \\
\begin{algorithm}[t!] 
\caption{Pseudo code for CLIP inversion.}
\textbf{Input}: Given source images and texts $(x_{s}, T_{s})$ \\
Given target text $T_{t}$ \\
Morphing images $x_{morph}$ are initialized as $x_{s}$ \\
\textbf{Output}: \( x_{\text{morph}} \) morphed by text guidance
\begin{algorithmic}
\STATE Obtain \( \Delta z^{T} = \frac{E_{T}(T_{t}) - E_{T}(T_{s})}{\left| E_{T}(T_{t}) - E_{T}(T_{s}) \right|} \) \\
\WHILE{\( t \) \textbf{in} \( \text{epochs} \)}
{  
    \STATE Detach gradient and re-initialize \( x_{\text{morph}} \) \\
    \STATE Apply \( N \) random augmentations to \( x_{\text{morph}} \) \\
    \STATE Get ensemble images of \( x^{Ens}_{\text{morph}} \) \\
    \STATE Calculate \( \Delta z^{I_{t}} = \frac{E_{I}(x^{Ens}_{\text{morph}}) - NE_{I}(x_{s})}{\left| E_{I}(x^{Ens}_{\text{morph}}) - NE_{I}(x_{s}) \right|} \) \\
    \STATE Minimize \( \frac{1}{N} \sum \mathcal{L}(\Delta z^{I_{t}}, N \Delta z^{T}) \) \\
    \STATE Update \( x_{\text{morph}} \)
}
\ENDWHILE
\STATE Obtain \( x_{\text{morph}} \)
\label{alg:clipinv}
\end{algorithmic}
\end{algorithm}

\begin{figure*}[ht!]
    \includegraphics[width=\linewidth]{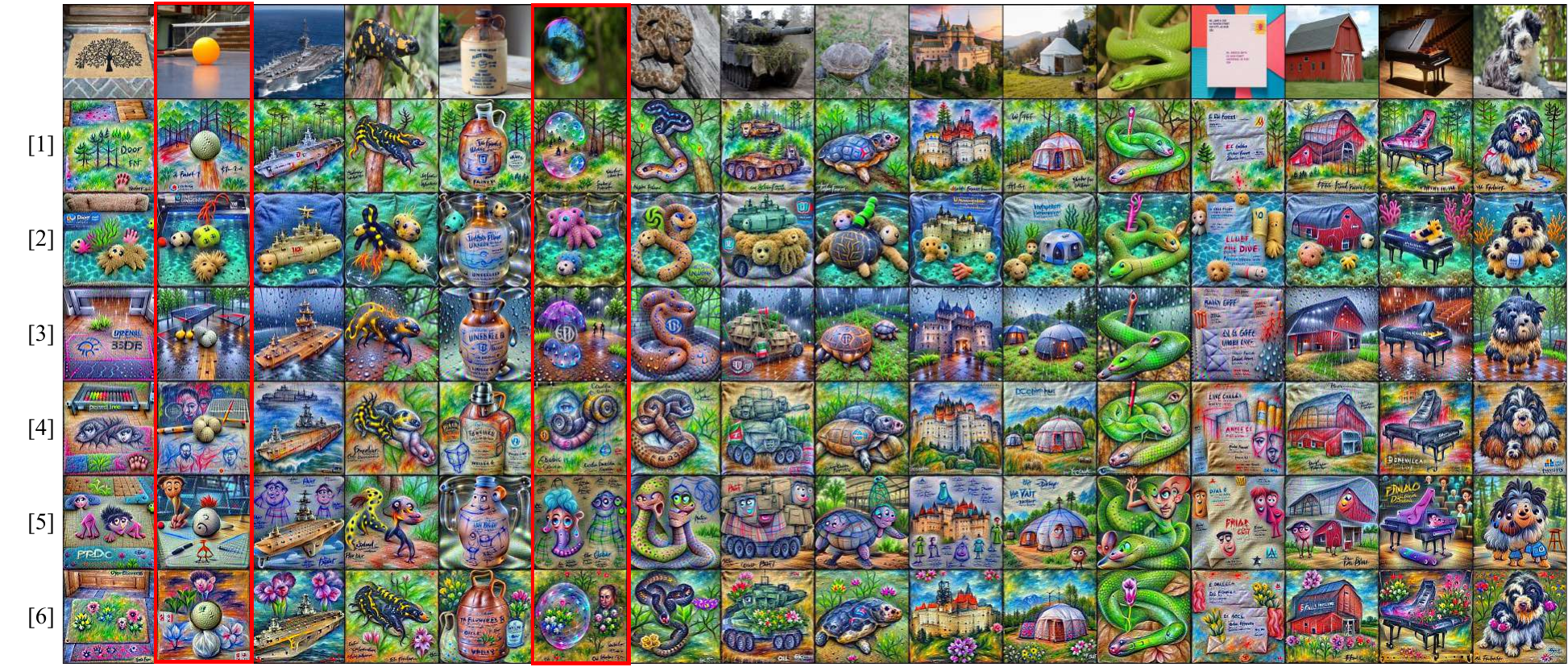}  \caption{Entire results of CLIP inversion using a directional CLIP loss to morph the given source images.}
    \label{fig:cos_inversion}
\end{figure*}
\begin{figure*}[ht!]
    \includegraphics[width=\linewidth]{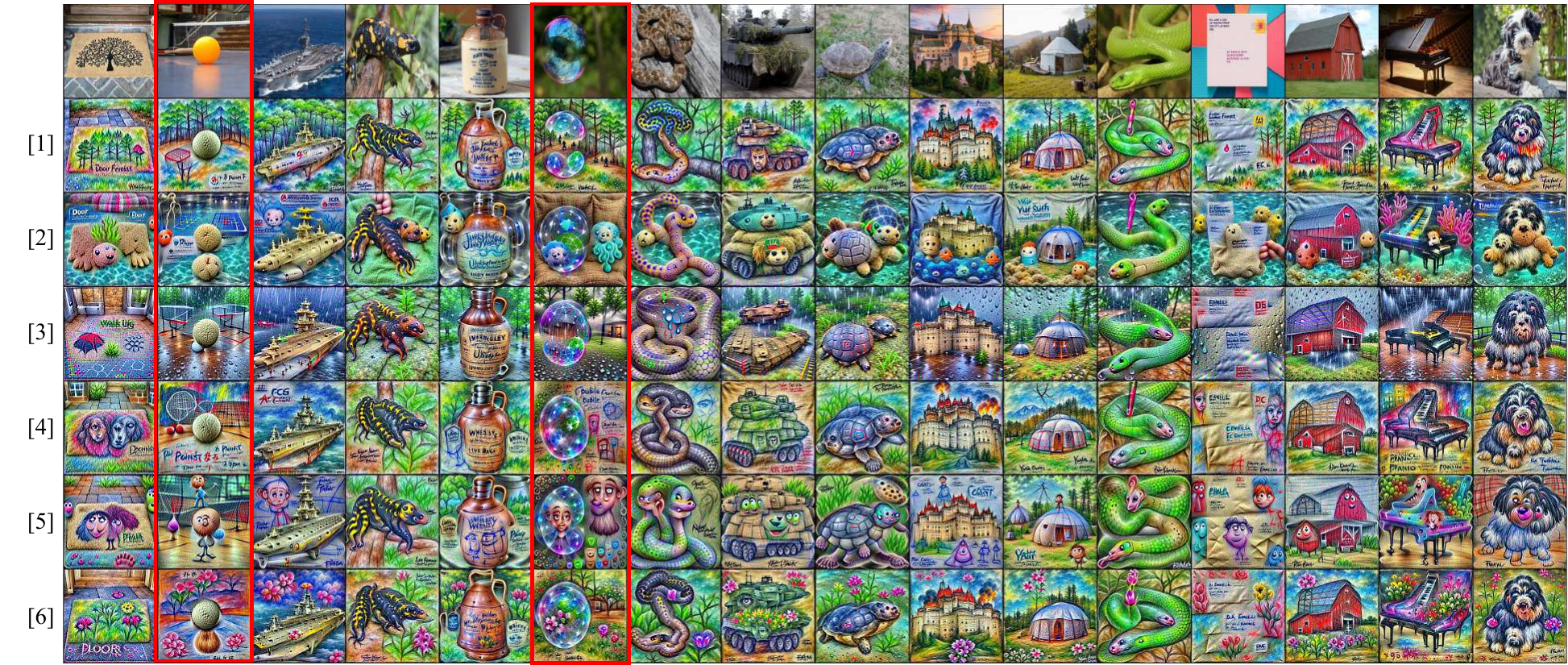}  \caption{Entire results of CLIP inversion using our proposed method to morph the given source images.}
    \label{fig:geo_inversion}
\end{figure*}
\begin{figure*}[t!]
    \includegraphics[width=\linewidth]{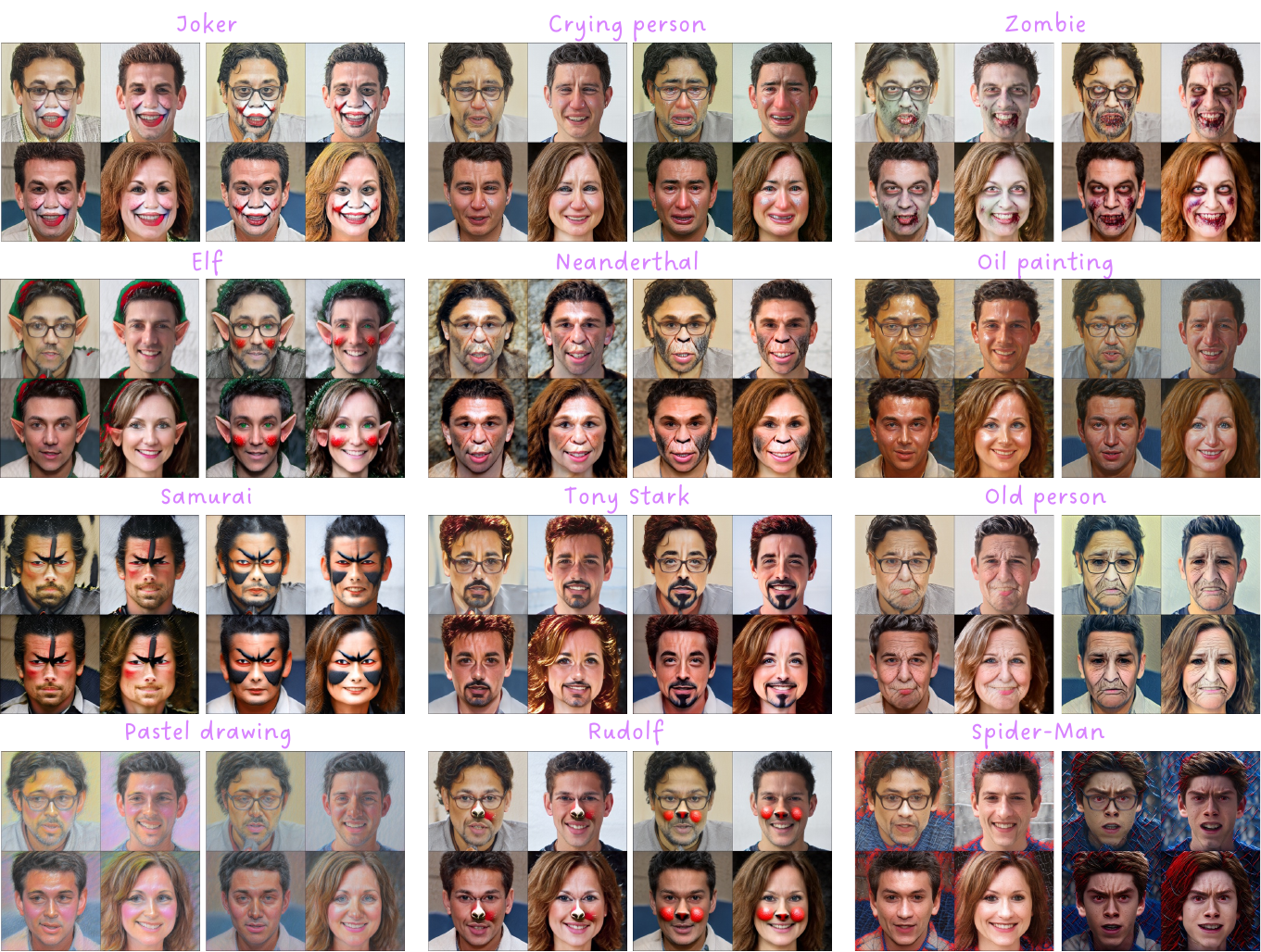}  \caption{Results of the CLIP-guided image morphing by StyleGAN-NADA with StyleGAN pre-trained on FFHQ dataset. In each plot, the left is the result of the baseline method, and the right is the result of our proposed method.}
    \label{fig:style_2}
\end{figure*} 
\indent We randomly selected $16$ classes from the ImageNet \cite{deng2009imagenet} dataset and randomly initialized the source images using real images obtained from the web.  Further, although we utilized real images as the source images for our CLIP inversion experiments, random noise can also be employed when no specific source images are available. In Fig. \ref{fig:cos_inversion} and \ref{fig:geo_inversion}, we present the entire results for the given classes, highlighting the significantly different outcomes with red boxes. Here, as described in previous model inversion literature \cite{yin2020dreaming,ghiasi2022plug,frans2022clipdraw}, the generated samples may not be visually plausible and photorealistic. 
\begin{figure*}[t!]
    \centering
    \includegraphics[width=0.7\linewidth]{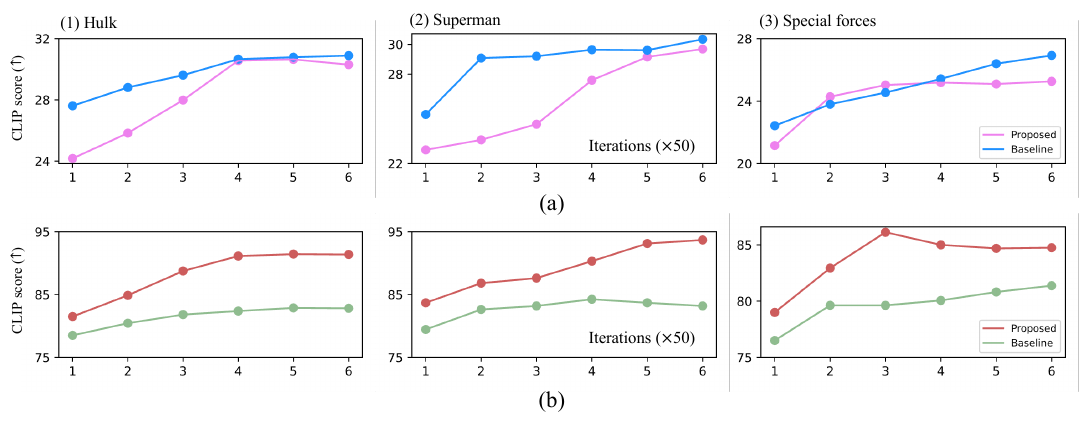} 
    \caption{CLIP scores of morphed images with the target texts (a) without modulating the modality gap distance, and (b) with modulating the modality gap distance.}
    \label{fig:clip_score_img2text}
\end{figure*}
\begin{figure*}
    \includegraphics[width=\linewidth]{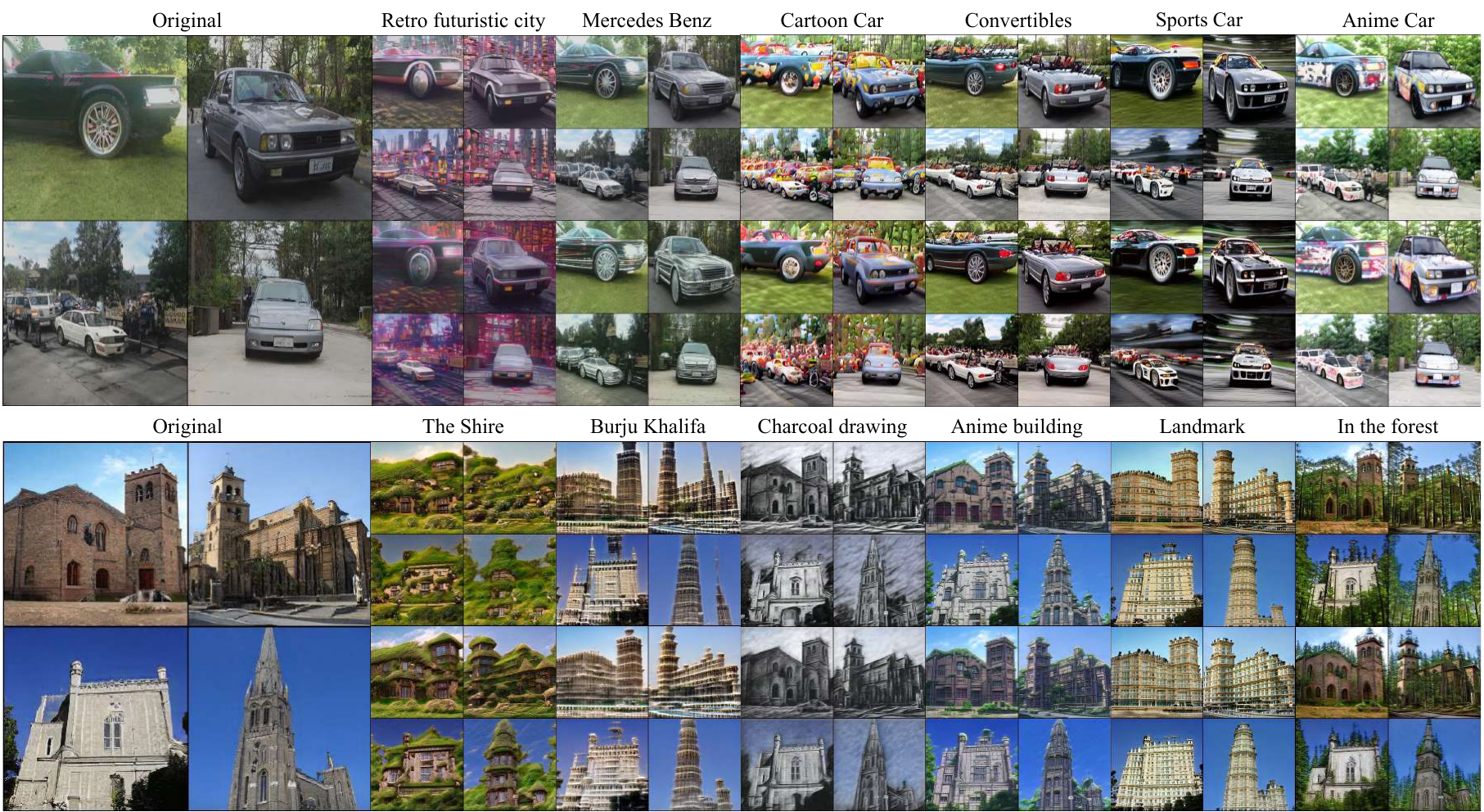} \caption{Results of the CLIP-guided image morphing by StyleGAN-NADA. Original images are generated from StyleGAN pre-trained with Stanford car and LSUN church datasets. In each plot, the upper row is the results of the baseline method, and the second row is the results of our method.}
    \label{fig:style_3}
\end{figure*}
\begin{figure*}
    \includegraphics[width=\linewidth]{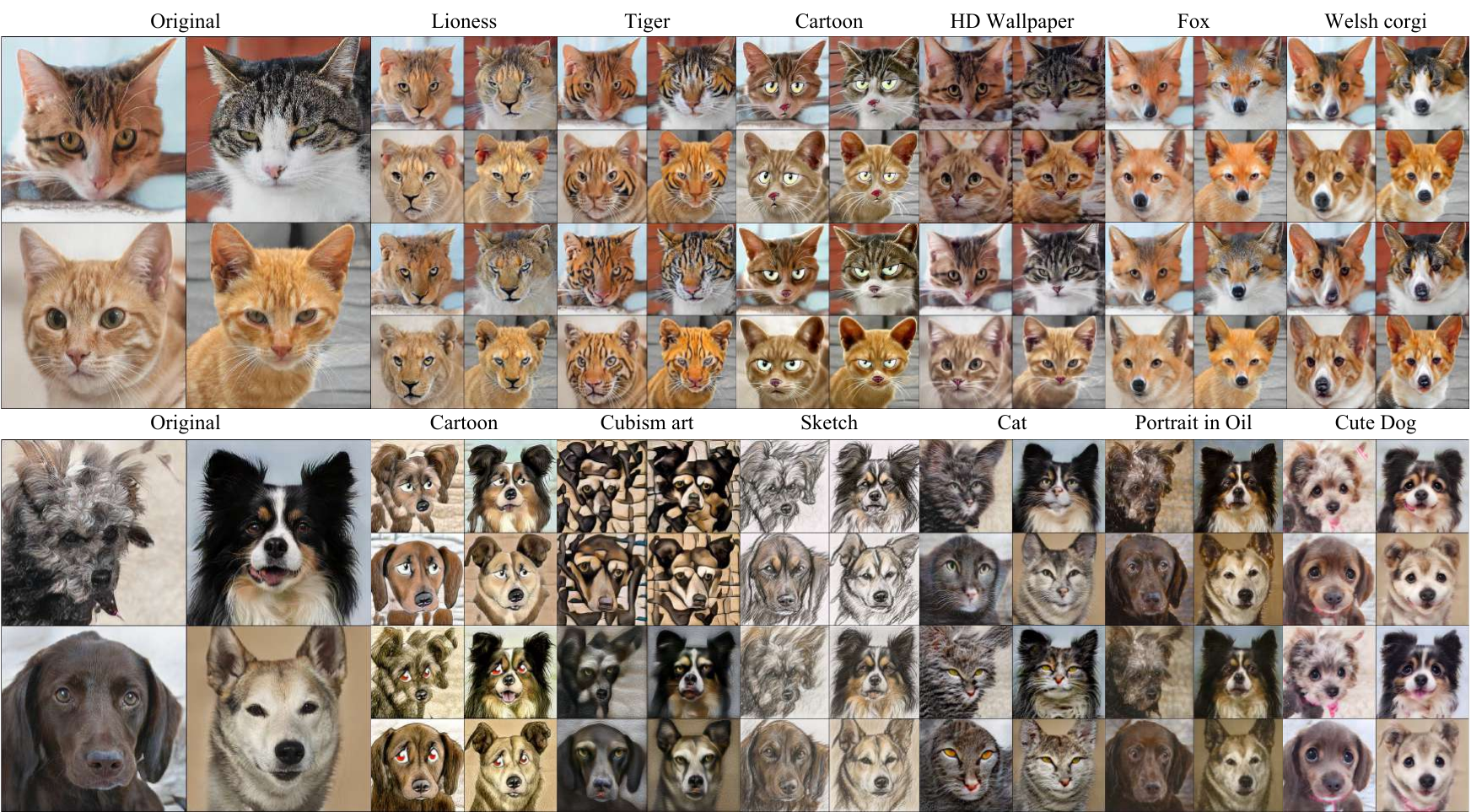} \caption{Results of the CLIP-guided image morphing by StyleGAN-NADA. Original images are generated from StyleGAN pre-trained with AFHQ cat and dog dataset. The upper row is the results of the baseline method, and the second row is the results of our method.}
    \label{fig:style_4}
\end{figure*}
\begin{figure}[t!]
    \includegraphics[width=\linewidth]{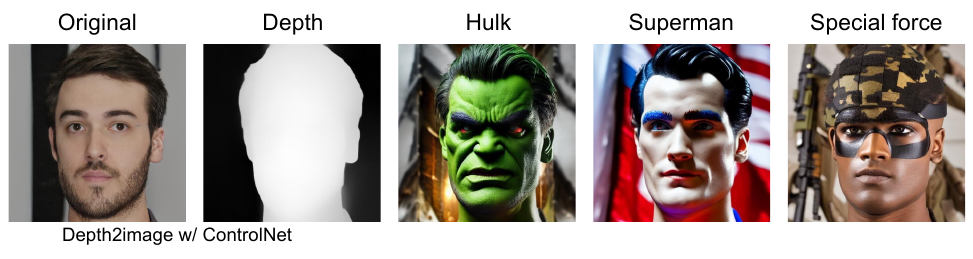} 
    \caption{ControlNet-generated examples for the prompts of Hulk, Superman, and Special Forces, which are used for the target images. To generate the images, the depth maps of source images are used.}
    \label{fig:controlNet}
\end{figure}
\section{Additional Results}
In this section, we describe additional implementation details and results for our experiments. 
\subsection{Further definitions}
\begin{definition}[Grassmann manifold]
For $0<N\leq D$, the space of $D\times N$ matrices with orthonormal columns is a Riemannian manifold, which constitutes the Stiefel manifold $St(N,D)$ \cite{harandi2015extrinsic} as expressed as follows.
\begin{equation}
        St(N,D)\triangleq{\{X\in R^{D\times N}: P^{T}P=I_{D}\}}.
    \label{eq:grassmann}
\end{equation}
\end{definition}
Here, all points on $St(N,D)$ that span the same subspace are grouped on the Grassmann manifold $\mathcal{G}(N,D)$ \cite{edelman1998geometry}. Here, $N$ and $D$ denote the batch size, and subspace dimension, respectively. 
\subsection{Time and memory consumption for calculating geodesic distillation loss}
The consuming time to obtain total loss is as follows. First, calculating SVD algorithm \cite{van1976generalizing} requires the costs of $\mathcal{O}(N^{2}d)$. Next, obtaining the geodesic ﬂow requires the costs of $\mathcal{O}(Nd)$, where $N$ denotes the batch size and $d$ denotes the dimension of subspace. As a limitation of our method, computational time linearly increases according to the subspace dimension to obtain SVD and geodesic flow. Further, if the given batch size is too large, the time complexity quadratically increases to obtain SVD and linearly increases to obtain geodesic flow, respectively. For future work, we plan to reduce the computational cost of our method. For computational resources, we used two A40 GPUs for CLIP inversion and a single A40 GPU for other CLIP-guided image morphing benchmarks in our experiment. Note that, as CLIP inversion uses $16$ ensemble images to morph one image at each iteration step, GPU memory requires more than 1GB for the batch size of $16$ while training.
\begin{table*}[t!]
    \centering
    \scalebox{0.75}{
    \centering
    \begin{tabular}{l|c|c|c|c|c|c|c|c|c}
        & \multicolumn{3}{c|}{LPIPS($\uparrow$)} & \multicolumn{3}{c|}{PSNR($\downarrow$)} & \multicolumn{3}{c}{SSIM($\downarrow$)} \\ \hline
        Prompts & Base & Ours & GT & Base & Ours & GT & Base & Ours & GT  \\ \hline
        Hulk & 0.20 $\pm$ 0.03 & 0.30 $\pm$ 0.03 & \textbf{0.65} $\pm$ \textbf{0.06} & 64.20 $\pm$ 1.34 & 64.27 $\pm$ 1.34 & \textbf{57.10} $\pm$ \textbf{1.17} & 0.46 $\pm$ 0.07 & 0.38 $\pm$ 0.07 & \textbf{0.24} $\pm$ \textbf{0.07}\\ \hline
        Superman & 0.32 $\pm$ 0.04 & 0.47 $\pm$ 0.04 & \textbf{0.59} $\pm$ \textbf{0.08} & 60.80 $\pm$ 1.28 & 59.90 $\pm$ 1.21 & \textbf{57.60} $\pm$ \textbf{1.16} & 0.39 $\pm$ 0.07 & \textbf{0.21} $\pm$ \textbf{0.04} & 0.31 $\pm$ 0.08\\ \hline
        Special force & 0.32 $\pm$ 0.04 & 0.32 $\pm$ 0.03 & \textbf{0.56} $\pm$ \textbf{0.07} & 62.10 $\pm$ 1.84 & 60.95 $\pm$ 1.72 & \textbf{58.80} $\pm$ \textbf{1.31} & 0.37 $\pm$ 0.05 & \textbf{0.32} $\pm$ \textbf{0.04} & 0.35 $\pm$ 0.08\\ 
    \end{tabular}
    }
    \caption{Measured scores of LPIPS, PSNR, and SSIM. Our method showed better results than the baseline method.}
    \label{tab:more_score}
\end{table*}
\begin{table*}[t!]
    \centering
    \scalebox{0.9}
    {
    \centering
    \begin{tabular}{l|c|c|c|c|c|c}
        \multirow{2}{*}{User study} & \multicolumn{2}{c|}{Hulk} & \multicolumn{2}{c|}{Superman} & \multicolumn{2}{c}{Special forces} \\ \cline{2-7}
         & Base & Ours & Base & Ours & Base & Ours \\ \hline
        Text-aligned ($\uparrow$) & 2.66 $\pm$ 0.97 & \textbf{3.25 $\pm$ 1.09} & 3.41 $\pm$ 0.94 & \textbf{3.74 $\pm$ 1.11} & 2.98 $\pm$ 1.16 & \textbf{3.43 $\pm$ 1.03} \\ \hline
        Photorealistic ($\uparrow$) & \textbf{2.97 $\pm$ 1.15} & 2.95 $\pm$ 1.34 & 3.58 $\pm$ 0.88 & \textbf{3.63 $\pm$ 1.04} & 2.85 $\pm$ 1.12 & \textbf{3.44 $\pm$ 1.10} \\ \hline
        Morphing Quality ($\uparrow$) & 3.12 $\pm$ 1.11 & \textbf{3.23 $\pm$ 1.19} & \textbf{3.42 $\pm$ 1.03} & 3.35 $\pm$ 1.07 & 2.95 $\pm$ 1.03 & \textbf{3.45 $\pm$ 1.11} \\
    \end{tabular}
    }
    \caption{User study to evaluate the quality of morphed images.}
    \label{tab:user_study}
\end{table*}
\begin{table}[t!]
    \scriptsize
    \scalebox{1.1}{
    \begin{tabular}{c|l}
        \multicolumn{1}{c}{} \\ \hline
            & 1) overall visual quality is real-looking. \\ Text-aligned  & 2) how well-applied given texts. \\ & 3) how well-resembled with given target images. \\ \hline
            & 1) overall visual quality is real-looking. \\ Photorealism & 2) how well-edited with visually plausible. \\ & 3) how well-resembled with given target images.  \\ \hline
            & 1) overall visual quality is real-looking. \\ Morphing & 2) how well-edited without losing the key attributes. \\ quality & 3) how well-resembled with given target images. \\ \hline
    \end{tabular}
    }
    \caption{The survey form used in our user study.}
    \label{tab:survey_table}
\end{table}
\section{Further results of benchmark studies}
\subsection{Additional results of StyleGAN-NADA}
In Fig. \ref{fig:style_2}, more results with various target texts of the baseline method and our proposed method are shown. Except for the target cases are `oil painting' and `pastel drawing' in which the source texts are 'photo', the source text is fixed as `human'. Here, considering together for all given target texts, our proposed method shows significantly better attribution-preserved morphing while successfully morphing the source images by following the text guidance. \\
\indent  Next, in Fig. \ref{fig:style_3}, additional results of morphed images from the Stanford cars \cite{KrauseStarkDengFei-Fei_3DRR2013} and LSUN churches \cite{yu2015lsun} are shown. For the Stanford car and LSUN church dataset, the resolution of the pre-trained generator is $512$. For both cases, our proposed method effectively morphs the given source images. For the car dataset, the used source text is set to `car', and the used target texts are `Retro-futuristic city', `Mercedes Benz', `Cartoon car', `Convertibles', `Sports car', and `Anime car'. Next, for the church dataset, except for the `Charcoal drawing' case in which the source text is `photo', the source text is fixed as `church'. The used target texts are `The Shier', `Burju Khalifa', `Charcoal drawing', `Anime building', `Landmark', and `In the forest'. Here, we fixed the random seed as $0$, iterations as $300$, and adaptive k as $10$, for all reproducing results. The style mixing coefficient of $0.9$ is used for `The Shier' and `Burju Khalifa', and otherwise uses $0$. \\
\indent In Fig. \ref{fig:style_4}, results of morphed images from the cats and dogs are shown. Note, for the AFHQ cat and dog dataset  \cite{choi2020stargan}, the resolution of the pre-trained generator is $512$. For both cases, our proposed method effectively morphs the given generated source images for the target texts. For the AFHQ cat dataset, except for the `HD Wallpaper' case in which the source text is `photo', the source text is fixed as `cat'. The used target texts are 'Lioness', `Tiger', 'Cartoon', 'HD Wallpaper', 'Fox', and `Welsh corgi'. Next, for the AFHQ dog dataset, except for the `Cat' case in which the source texts are 'dog', the source text is fixed as `photo'. The used target texts are `Cartoon', `Cubism art', `Sketch', `Cat', `Portrait in oil', and `Cute dog'. Here, we fixed the random seed as $2$, and the style mixing coefficient of $0.9$ is used for `Convertibles' and `Sports car', and otherwise uses $0$. Other hyperparameters are the same as in the abovementioned case. 
\subsection{Additional evaluations}
\subsubsection{Metric scores}
\indent In our subsequent experiments, we measured the CLIP scores of the morphed images and target texts. Fig. \ref{fig:clip_score_img2text} illustrates the scores obtained from the baseline method and our proposed method. To evaluate the scores, images were generated with $100$ per prompt and iteration. Interestingly, although our method morphs the images to have significant changes from the source images and guides the images to approach close to the target image manifold with superior quality, our method consistently yielded lower CLIP scores in (a). In this case, (a) represents the results without modulating the modality gap distances. This intriguing observation indicates that the existing img2text CLIP scores, which measure the cosine similarity between image and text features, do not necessarily reflect the level of photorealism of the morphed images aligned with the target texts. On the contrary, in (b), which represents the results with modulated optimal modality gap distances, our proposed method outperforms the CLIP scores of the baseline method for all training iterations and prompts. Therefore, based on our investigations, we conjecture that modulating the modality gap distance is a crucial key for assessing the quality of morphed images with given texts. To the best of our knowledge, this finding is the first to measure the extent of alignment of quality of morphed images and texts in CLIP space while image morphing by modulating the modality gap distances. \\
\indent  To provide comprehensive evaluations, we measured LPIPS \cite{zhang2018unreasonable}, PSNR, and SSIM \cite{wang2004image} scores with source images. Table. \ref{tab:more_score} presents the results, demonstrating that our method consistently outperforms the baseline method across all prompts and metrics. Here, as shown in Fig. \ref{fig:controlNet}, target images are generated with ControlNet \cite{zhang2023adding}. As crawled the target images from the web are highly unrefined images, we created a target dataset for an elaborate comparison. A total of $500$ samples were generated per prompt. In the figures, `GT' represents the scores between target images and source images. 
\subsubsection{Human evaluations}
\indent In order to reduce the subjectivity involved in evaluating the quality of morphed images, we conducted several user studies. To prepare the data, we sampled data from checkpoints with the maximum CLIP scores and used it for user studies. We employed Amazon Mechanical Turk for user studies, providing detailed instructions. We conducted three different tests, each comprising 100 random samples per prompt. In each test, 100 individuals evaluated different pairs of images three times to determine which ones (\textit{i.e.} baseline and ours) exhibited superior qualities in terms of text alignment, photorealism, and qualitative morphing. Consequently, as described in Table \ref{tab:user_study}, our method demonstrated comparable or superior performance compared to the baseline method. The used survey forms in our user study are described in Table. \ref{tab:survey_table}.
\subsection{Additional results of Text2Live}
\begin{figure}[t!]
    \includegraphics[width=\linewidth]{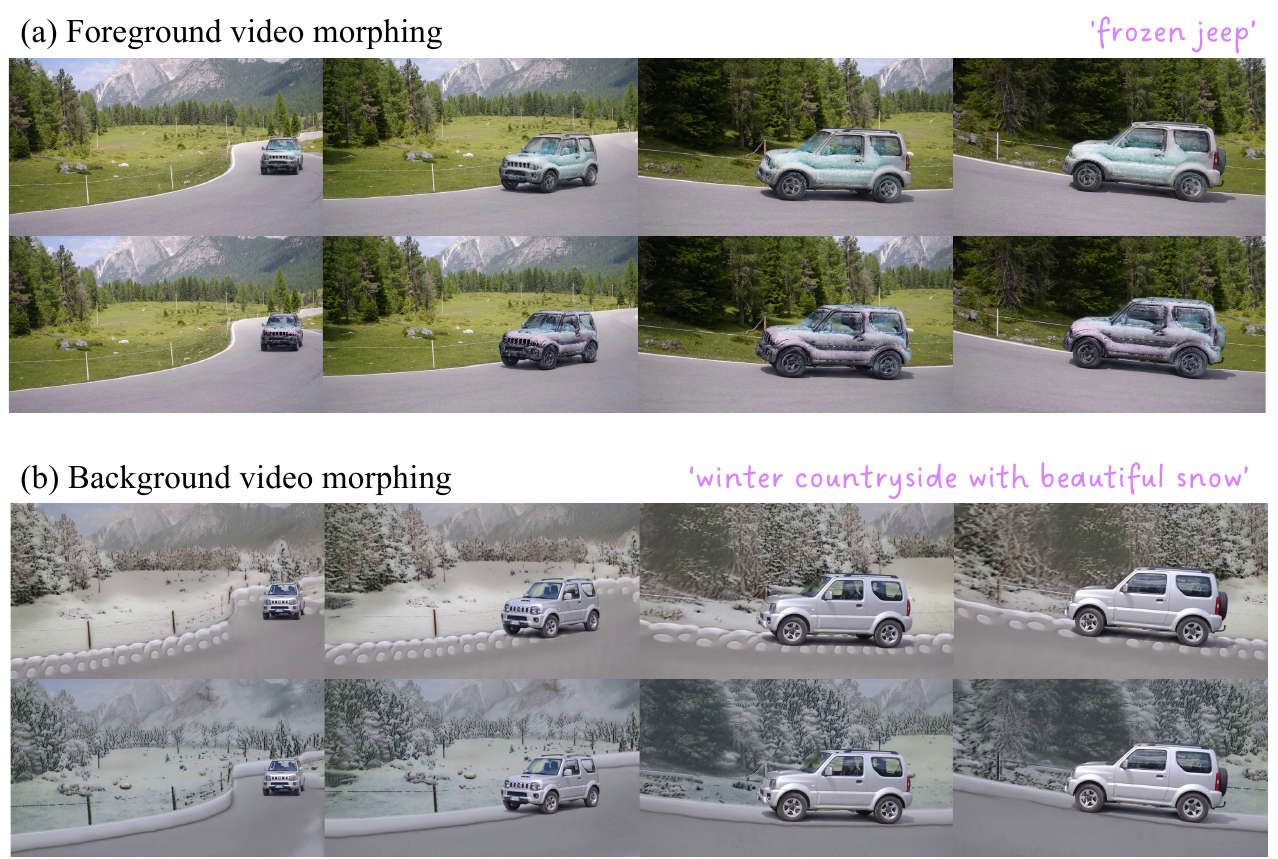}  \caption{Additional results of Text2Live for (a) foreground and (b) background video morphing. In this figure, the first row represents the results of the baseline method, and the second row represents the results of our proposed method.}
    \label{fig:suppl_textlive}
\end{figure}

For further experiments with the Text2Live benchmark, in Fig. \ref{fig:suppl_textlive}, foreground video morphing results with the given target texts of `frozen jeep' and `winter countryside with beautiful snow' are represented. For foreground video morphing, our proposed method showed better morphing results by reflecting the text guidance to transform the texture of given images. Next, for background video morphing, our proposed method showed better photorealistic video morphing results (\textit{i.e.}, the outline of the road is covered with snow) reflecting the given texts. Here, we used `car-turn' video frames provided by \cite{bar2022text2live}, the source text is set to `jeep', and fixed random seed as $1$ for both reproducing experiments of the baseline and our proposed method. \\
\begin{figure}[t!]
    \includegraphics[width=\linewidth]{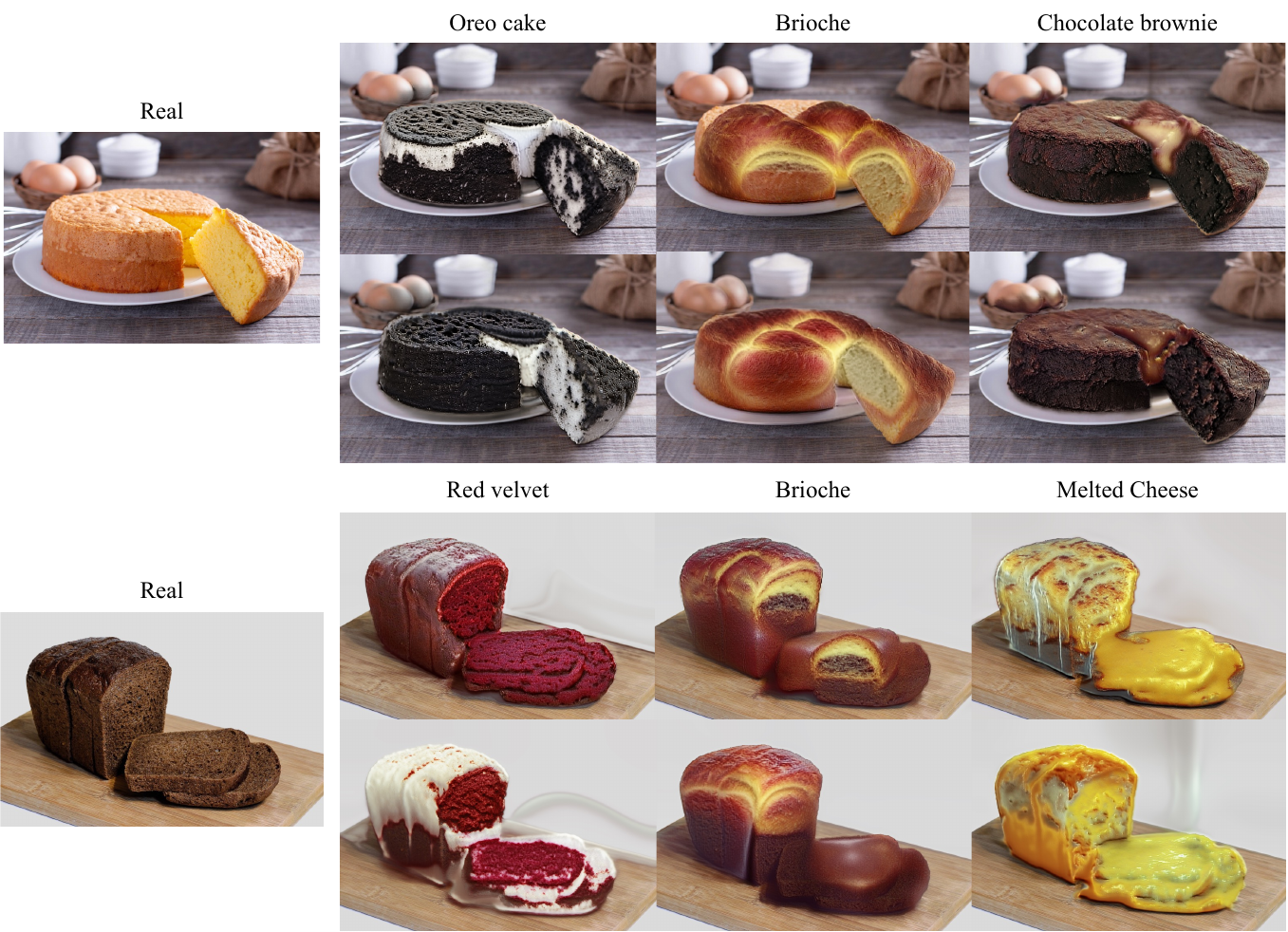} \caption{Two random source images, morphed images, and corresponding target texts are described. The first row is the results of the baseline method, and the second row is the results of our proposed method.}
    \label{fig:text2live_image}
\end{figure}
\indent Fig. \ref{fig:text2live_image} illustrates the morphing outcomes obtained from the provided real source images and selected target prompts for both the baseline method and our proposed method. The discrepancies were less noticeable for oreo cake and brioche. However, clear distinctions were observed in the case of chocolate brownies, red velvet, and melted cheese. 
\section{Further CLIP-guided image morphing benchmarks}
\subsection{Improving StyleCLIP}
\begin{figure}[h!]
    \includegraphics[width=\linewidth]{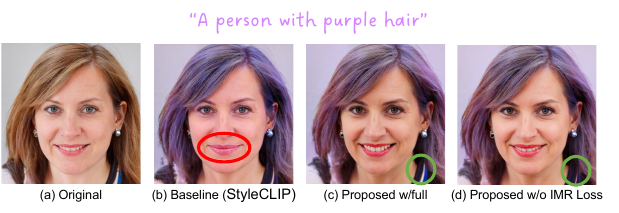}  \caption{Results of StyleCLIP. Note, although the baseline suffers from severe forgetting of detailed attributes, our method significantly enhanced morphing quality not hurting the important attributes.}
    \label{fig:styleclip}
\end{figure}

\indent StyleCLIP \cite{patashnik2021styleclip} proposed a text-guided latent optimization method, which aims to directly optimize a latent vector in StyleGAN generator’s $\mathcal{W+}$ space, with a cosine loss between the CLIP features of the given image and the text. For regularization, $L2$ and Arcface \cite{deng2019arcface} losses are used. Here, the used input $\mathcal{W+}$ vectors for real images are obtained from a pre-trained e4e \cite{tov2021designing} encoder. To validate the effectiveness of our method in real image editing scenarios with moderate prompts, we conducted a comparative analysis with the baseline method, as depicted in Figure \ref{fig:styleclip}. As a result, while the baseline method (b) produced morphed images with closed lips, our method (c) demonstrated superior preservation of attributes and achieved more photorealistic outcomes compared to the baseline method.
\begin{figure}[t!]
    \includegraphics[width=\linewidth]{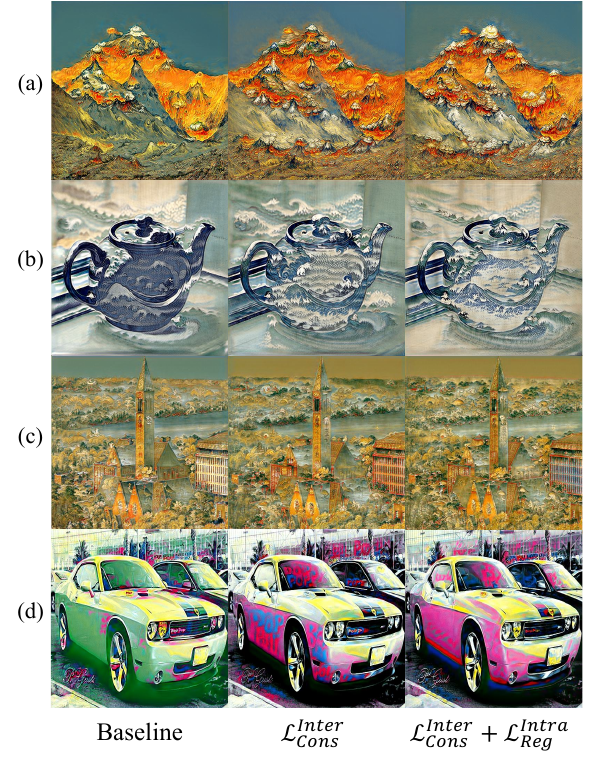}  \caption{Implementation results of CLIP-styler as a benchmark study. The used prompts are, $(a)$ `Volcano eruption with the style of Vincent Van Gogh', $(b)$ `The great wave off Kanagawa by Hokusai', $(c)$ `Oriental painting', and $(d)$ `Pop art'.}
    \label{fig:style_tr}
\end{figure}
\begin{figure}[t!]
    \includegraphics[width=\linewidth]{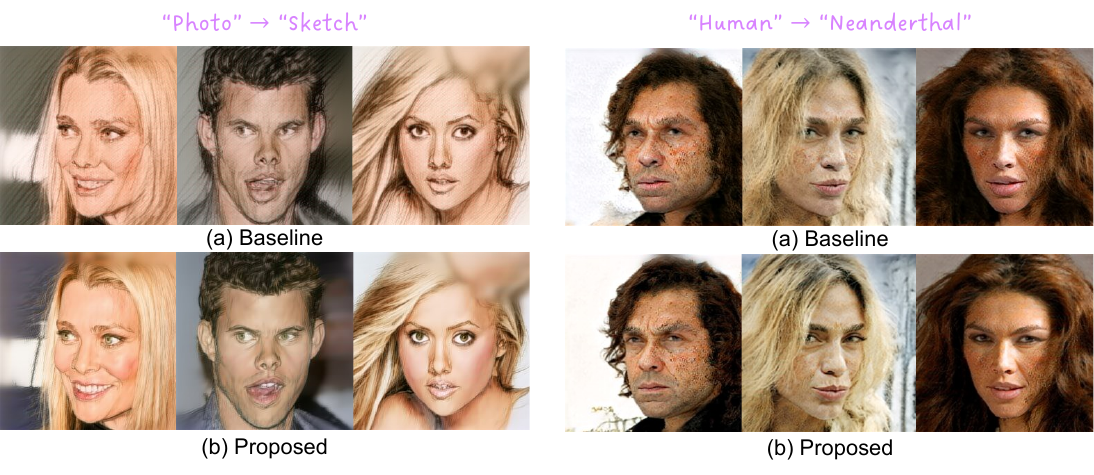}  \caption{Additional Results of DiffusionCLIP.}
    \label{fig:diffusionclip}
\end{figure}
\begin{figure*}[t!]
    \includegraphics[width=\linewidth]{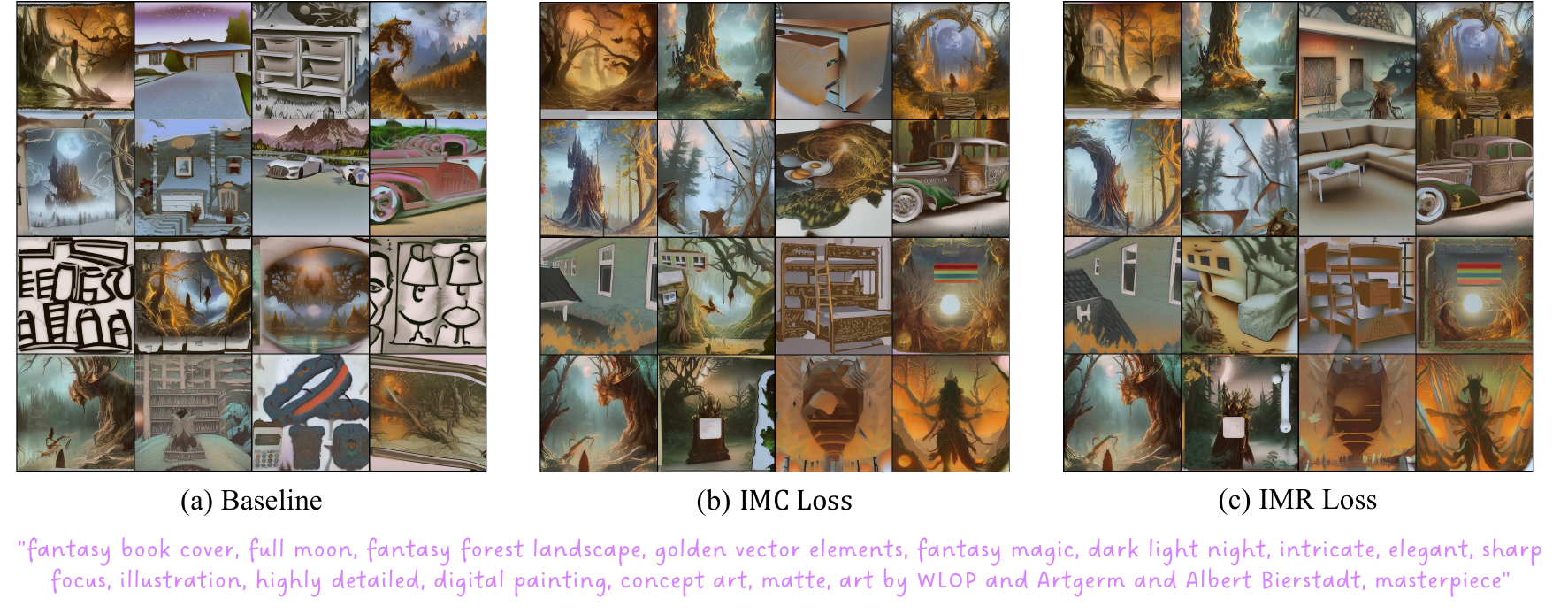}  \caption{Generated images by CLIP-guided Stable Diffusion from a given prompt. Here, images are generated via (a) the baseline method, (b) only considering IMC loss, and (c) our proposed loss (IMC+IMR).}
    \label{fig:diffusion_1}
\end{figure*}
\begin{figure*}[t!]
    \includegraphics[width=\linewidth]{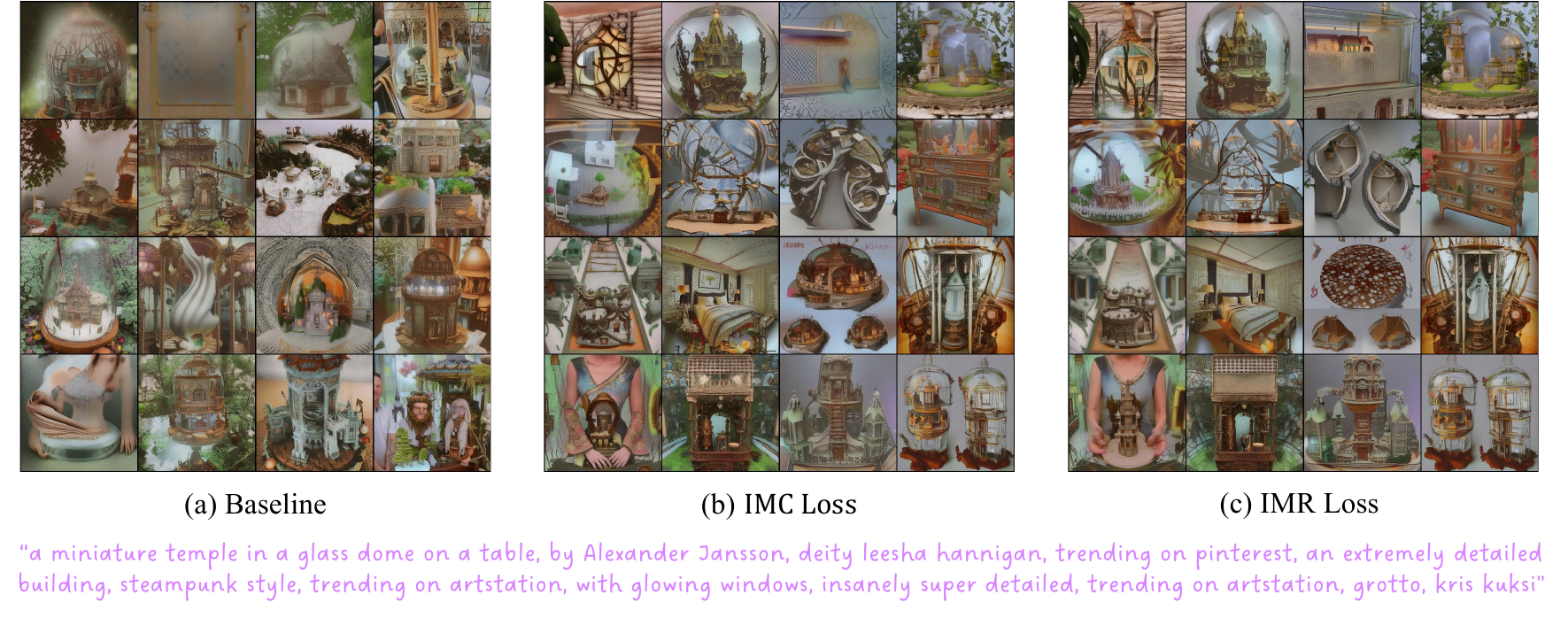}  \caption{Other generated images by CLIP-guided Stable Diffusion from a given prompt. Here, images are generated via (a) the baseline method, (b) only considering IMC loss, and (c) our proposed loss (IMC+IMR).}
    \label{fig:diffusion_2}
\end{figure*}
\subsection{Improving CLIP-guided Style Transfer}
Kwon \textit{et. al.} \cite{kwon2022clipstyler} proposed a new framework for the CLIP-based style transfer method that allows transferring style concepts to given content images without style images. While transforming the given content images, only a text description is needed for the desired concept to morph the images. To this end, they proposed patch-wise matching loss for text-image pairs, employed multi-view augmentations, and enforced rejection threshold for the value of cosine similarity. As a result, the given images could be morphed to get realistic textures by given query texts. \\
\indent To verify the effectiveness of our proposed method with the extent of CLIP-guided style transfer, in Fig. \ref{fig:style_tr}, we compared the baseline method and our proposed method for text-guided style transfer. The original images are provided by \cite{kwon2022clipstyler}. The used source prompts are fixed as `Photo', and target prompts for experiments are described in the figure. As a result, compared to CLIP-styler \cite{kwon2022clipstyler} baseline, by replacing the directional CLIP loss, the results showed the better-enhanced quality of morphed images. For instance, in the case of a prompt for `The great wave off Kanagawa by Hokusai' is applied on the teapot content image, the style transferred image by the baseline method showed relatively localized morphing results as the waves are not effectively covered on the source image. However, the morphed images show better results as semantically realistic morphing by replacing the original directional CLIP loss as IMC loss ($\mathcal{L}^{Inter}_{Cons}$). Next, by fully combining the proposed loss ($\mathcal{L}^{Inter}_{Cons} + \mathcal{L}^{Intra}_{Reg}$), the results showed more enhanced photorealistic morphing. Here, we note that we did not alter any used hyperparameters of \cite{kwon2022clipstyler}, and threshold rejection to prevent over-stylization of content images proposed in \cite{kwon2022clipstyler} is not used for minimizing our proposed loss. The given source image for the mountain is from the web, and teapot, church, and car images are brought from \cite{kwon2022clipstyler}. 
\subsection{Improving DiffusionCLIP}
\indent With another baseline \cite{kim2022diffusionclip}, in Fig. \ref{fig:diffusionclip}, we have observed that the results of the proposed method appear to be closer to the source domain (\textit{i.e.} `photo') than the results of the baseline method. Upon closer examination of the morphed images, in the case of the `Sketch' prompt represented in the left figure, the images generated using the proposed method reflect the target prompt while preserving the characteristics of the original source image. For the `Neanderthal' prompt in the right figure, we have observed that there is not a significant difference compared to the baseline method. Thus, to better verify the effectiveness of our proposed method, using extreme prompts result in more different outcomes. 
\subsection{Improving CLIP-guided Latent Diffusion Model}
In this section, to highlight the advantages of our proposed method, we describe the additional improvements of CLIP-guided latent diffusion model \cite{rombach2022high} by using diffusers python library. Note, compared to the CLIP-guided image morphing, a pre-trained text-to-image (TTI) diffusion model \cite{rombach2022high} could generate the image from the random noise without requiring the initialization of source images. However, with the aid of CLIP guidance which exploits CLIP's image and text encoders on CLIP space, the diffusion generator could generate more realistic images than using the generator alone, by guiding the generator at every denoising step. Note, the baseline method minimizes the spherical distance \cite{crowson2022vqgan} to acquire CLIP guidance to morph the generated images on CLIP space. In this section, as an ablation study, we present our proposed method's contributions to enhancing the existing CLIP-guided TTI generation methods. Here, for TTI generation, we used a stable diffusion v1-5 pre-trained model, ViT-B/32 CLIP text encoder, and VAE encoder and decoder models provided from the web. \\
\indent Here, in our experiments, we used specified prompts that are randomly selected from the web for TTI generation. In Fig. \ref{fig:diffusion_1} and Fig. \ref{fig:diffusion_2}, we used the batch size as $1$ for sampling and fixed the manual seed as $0$ for the random number generator. \\
\indent Next, OpenCLIP \cite{cherti2022reproducible} is used to guide the diffusion model in CLIP space, which model is trained with LAION-2B\footnote{LAION-2B is a subset of LAION-5B \cite{schuhmann2022laion} which contains 2.32 billion samples consisting of image and English text pairs.} dataset. Interestingly, the baseline method poorly results in both cases as represented in Fig. \ref{fig:diffusion_1} and \ref{fig:diffusion_2} except for several cases. In contrast, in the same environmental settings, our proposed method consistently outperforms the baseline method by generating more photorealistic images in almost all of the cases.
{\small

\bibliographystyle{named}
\bibliography{references}